\documentclass[final,1p,times,twocolumn,authoryear]{elsarticle}

\usepackage{amssymb}
\usepackage{amsthm}
\usepackage{amsmath}
\usepackage{color}
\usepackage{amsfonts} 
\usepackage{hyperref}

\begin{document}

\begin{frontmatter}



\title{Global Symmetry and Orthogonal Transformations from Geometrical Moment $n$-tuples}

\author[1]{Omar Tahri }
\ead{otahri@gmail.com}


\affiliation[1]{
	organization={ICB UMR CNRS 6303, Universite de Bourgogne},
	addressline={9 Av. Alain Savary},
	city={Dijon},
	postcode={21000},
    state={Bourgogne},
	country={France}
}
\begin{abstract}

Detecting symmetry is crucial for effective object grasping for several reasons. Recognizing symmetrical features or axes within an object helps in developing efficient grasp strategies, as grasping along these axes typically results in a more stable and balanced grip, thereby facilitating successful manipulation. This paper employs geometrical moments to identify symmetries and estimate orthogonal transformations, including rotations and mirror transformations, for objects centered at the frame origin. It provides distinctive metrics for detecting symmetries and estimating orthogonal transformations, encompassing rotations, reflections, and their combinations. A comprehensive methodology is developed to obtain these functions in n-dimensional space, specifically moment \( n \)-tuples. Extensive validation tests are conducted on both 2D and 3D objects to ensure the robustness and reliability of the proposed approach. The proposed method is also compared to state-of-the-art work using iterative optimization for detecting multiple planes of symmetry. The results indicate that combining our method with the iterative one yields satisfactory outcomes in terms of the number of symmetry planes detected and computation time.
\end{abstract}




\end{frontmatter}


\section{Introduction}
\label{sec:introduction}

The problem of extracting geometrical features holds a prominent position within the literature of pattern recognition. 
Specifically, the identification of characteristics such as symmetry or orthogonal transformations remains pertinent across various domains.  
For instance, symmetrical properties find utility in fields such as chemistry, where they are leveraged in the design of expansive protein domains~\cite{fortenberry2011exploring}.
Additionally, in structural engineering, the discovery of symmetries within polygonal structures serves as a significant aspect~\cite{zhang2022structural}
These symmetries are also integral to structure-aware processing, as highlighted by~\cite{mitra2014structure}, as they constitute a fundamental source for generating structural patterns.
Moreover, within the computational domain, symmetry detection emerges as a critical feature, notably in constraint programming~\cite{gent2006symmetry} and linear programming.
Such detection aids in problem simplification, as emphasized by~\cite{margot2009symmetry}.

In general, the search for a robust model capable of detecting symmetry holds significance across various domains such as computer graphics, computer vision, machine learning~\cite{liu2010computational}, and robotics.
For instance, in the domain of object pose estimation, symmetry properties have been effectively utilized to identify matching correspondences, as demonstrated by~\cite{song2020hybridpose}.
Additionally,~\cite{huang2014image} proposed a method leveraging symmetries for automatic image completion, highlighting the broad applicability of symmetry detection techniques. 
Furthermore, applications including object completion~\cite{speciale2016symmetry} and model compression~\cite{tayangkanon20183d,wang2015progressive} also stand to benefit from the exploitation of symmetries and orthogonal transformations.

Robotics field also exploits these features for extracting object properties.
This is fundamental to predict their behavior when they are in motion.
For instance, detecting objects symmetries will contribute developing better grasping and manipulation strategies~\cite{lederman2003perceptual,li2005fast}.

The literature encompasses a multitude of specialized methodologies owing to the extensive range of applications. 
For instance, certain researchers, such as those in~\cite{atallah1985symmetry, wolter1985optimal}, focus on estimating perfect symmetry, while others, as~\cite{alt1988congruence}, address the estimation of approximate global symmetry within discrete sets of points. 
Addressing the 2D scenario,~\cite{derrode2004shape} employ the Fourier-Mellin transform to assess motion parameters encompassing rigid transformations and reflection symmetries.
Alternatively, some researchers opt for feature matching techniques to estimate rotational and reflection symmetries in 2D images, as demonstrated in~\cite{shen1999symmetry} employing features derived from the first three nonzero generalized complex (GC) moments. 
Additionally, the Slope Chain approach has garnered attention in the literature~\cite{aguilar2020detection}, although its efficacy is compromised when dealing with quasi-symmetrical distributions. 
Proposing a distinct methodology, \cite{wang2015reflection} present a method for detecting reflection symmetry via the correspondence of locally affine invariant edge-based features. 
In another vein, \cite{widynski2014local} advocate for utilizing contours of ribbon-like 2D objects within 2D images to discern smooth local symmetries. 
Furthermore, recent advancements have seen the emergence of neural network-based approaches for symmetry detection, as exemplified by the work of~\cite{krippendorf2020detecting}.

In three-dimensional (3D) space, \cite{ecins2017detecting} propose employing symmetrical fitting as a methodology for designing a reflection symmetry detection technique. 
Additionally, methods have been introduced for detecting global reflection in 3D point clouds~\cite{nagar20203dsymm}. 
\cite{ji2019fast} advocate for a neural network-based approach aimed at detecting 3D reflection symmetry planes through pointwise classification.
Continuing along this idea, \cite{zhou2021nerd} propose a neural network architecture utilizing ResNet as its backbone.
Conversely, \cite{gao2020prs} leverage a convolutional neural network (CNN), with emphasis placed on refining the loss function. 
Other approaches, such as that proposed by \cite{shi2020symmetrynet}, rely on numerical optimization techniques, leveraging symmetry priors to address reconstruction challenges in noisy input scenarios.
Alternatively, closed-form methodologies are also evident in the literature, exploiting various geometrical features. 
For instance, 3D moments have been utilized to deduce symmetry parameters, as evidenced by \cite{martinet2006accurate}. 
Over the span of more than half a century, theoretical frameworks and applications leveraging moments have thrived across diverse domains, encompassing tasks such as pose estimation \cite{cyganski1985applications, mukundan1998moment}, character recognition \cite{wong1995generation}, target recognition \cite{liu2012target}, quality inspection \cite{sluzek1995identification}, image matching \cite{chen2010zernike}, multi-sensor fusion \cite{markandey1992robot}, and visual servoing \cite{tahri2005point, tahri2010decoupled}.

The literature also features works that tackle the challenge of symmetry detection in high-dimensional spaces by projecting the problem onto lower-dimensional spaces. 
Examples include studies such as \cite{zhang2022structural, chen2017computational, zingoni2012symmetry}, which identify symmetries within intricate 3D topologies by projecting them onto 2D images. 
In contrast, our research introduces a comprehensive approach to detecting symmetrical and orthogonal transformations (encompassing rotation and mirror transformations) without the need for mapping input data to lower-dimensional spaces.
We present a closed-form solution based on moments that can be applied across various domains regardless of their space dimensionality. 
Furthermore, as an analytical solution, our approach does not require prior data for model training.

Subsequent to this introduction, we will provide an overview of the framework and its adaptation to each specific problem.
Subsequently, to assess the $n$-dimensional condition, we will conduct a comprehensive analysis across multiple levels, encompassing both 2D and 3D perspectives. 
Furthermore, the evaluation will be conducted with respect to symmetry patterns, including reflection planes, axes of rotation, or centers of gravity.

\section{Geometrical representation using moments}
Let define first the fundamentals of the geometrical features used here and how they variate with respect to the motion.
Concretely, this work considers the manifold of \emph{p}-th order moments of an original \emph{n}-dimensional space. 

\subsection{Geometrical p-th order moments}
Consider, a defined region ${\mathbf{X}}\subset\mathbb{R}^n$ by a piecewise-continuous real density distribution function $f({\mathbf{X}})$ of nonzero values only in a finite region of the original $n$-dimensional space.
Its moments of order $p=\sum_{i=1}^{n}p_i$  is defined by,
\begin{equation}
	\label{MomentsND}
		m_{p_1 \cdots p_n}=\int_{\mathbb{R}^n}\prod_{i=1}^{n} x_{i}^{p_i}f\left({\bf X}\right) d{\bf X},
\end{equation}

\noindent where $\left[x_1  \cdots x_n\right]^{\top}$ are the components of an infinitesimal point $\mathbf{x}\in\mathbf{X}$. The central moments of order $p$ are defined by,
\begin{equation*}
	\label{MomentsNDCentred}
	\mu_{p_1 \cdots  p_n} = \int_{\mathbb{R}^n}\prod_{i=1}^{n} \left(x_{i}-\overline{x}_{i}\right)^{p_i}f\left({\bf X}\right) d{\bf X},
\end{equation*}

\noindent where each $\overline{x}_{i}$ is a component of the object gravity center $\overline{\mathbf{x}} = \left[\frac{m_{1,0 \cdots  0}}{m_{0,0 \cdots  0}}\quad\frac{m_{0,1,0\cdots  0}}{m_{0,0 \cdots  0}}\quad\cdots\quad\frac{m_{0 \cdots 0,1}}{m_{0,0 \cdots  0}}\right]^{\top}$ .
Moving the spatial reference frame to this point allows computing centered moments. 
In the case  when the generative function $f(\mathbf{X})$ defines a discrete distribution function, their moments are defined by,
\begin{equation}
	\label{MomentsNDdiscrete}
	m_{p_1 \cdots p_n} = \sum_{i=1}^{\left|\mathbf{X}\right|} \prod_{j=1}^{n} {x_i}_{j}^{p_j}f\left({\bf x}_i\right)
\end{equation}

\noindent where $\mathbf{x}_i$ is a point of the point set $\mathbf{X}$ and  ${\left|\cdot\right|}$ gets the amount of points.
The proposed framework in this paper is valid for both continuous and discrete objects, namely a discrete 3D points cloud.

\subsection{Orthogonal transformation and  rotational speed}
The motion of a distribution $\mathbf{X}$ under $n$-dimensional orthogonal transformation ${\bf R}\in$~SO~($n$) is expressed as follows,
\begin{equation}
	\label{Rotations}
	{\bf X'}={\bf R}{\bf X}.
\end{equation}
\noindent satisfying ${\bf R}{\bf R}^\top={\mathbf{\mathbb{I}}}$ and det~$\left({\bf R}\right)=1$. If a rotational speed is applied, then the speed of each point ${\dot{\bf X}}$ is given by,
\begin{equation}
	\label{Rotationsvar}
	{\dot{\bf X}}={\bf L}{\bf X}
\end{equation}
\noindent where ${\bf L}$ is an antisymmetric matrix defined from  rotational velocities.
Particularly, we have $\dot{\bf x}={\bf L_{3}}{\bf x}=-\left[  {\bf x}  \right]_{\times} {\boldsymbol\omega} $ and a defined matrix ${\bf L_{3}}$ using the vector of values ${\boldsymbol\omega} =\left[\omega_1\quad\omega_2\quad\omega_3\right]^{\top}$ of rotation speeds in 3D distributions $\mathbf{X}\subset\mathbb{R}^3$ for the $x_2x_3$, $x_3x_1$ and $x_1x_2$ planes. Concretely, the antisymmetric defined matrix is defined as follows,
\begin{equation}
  \label{A3D}
  {\bf L_{3}}=
  \left[
      \begin{array}{ccc}
          0         & -\omega_3 & \omega_2  \\
          \omega_3  & 0         & -\omega_1 \\
          -\omega_2 & \omega_1  & 0
      \end{array}
      \right]
\end{equation}

\noindent we derive that both eq.~\ref{Rotationreal} and eq.~\ref{Rotations} are linearly dependent with respect to the same orthogonal transformation $\mathbf{R}$.

\subsection{Moment time variation and rotational speeds}
Considering that an orthogonal transformation in $\mathbf{X}$, eq.~(\ref{Rotations}), implies motion in its moments.
The motion, or time variation, of a moment $ m_{p_1 \cdots p_n}$~\cite{tahri2004utilisation,Tahri03b} is obtained by differentiating eq.~(\ref{MomentsND}),
\begin{equation}
	\label{dmp1ppn}
    \begin{array}{l}
		\dot{m}_{p_1 \cdots  p_n}=\sum_{i=1}^{n}\int_{\mathbb{R}^n} p_i\dot{x}_i x_i^{p_{i}-1}\prod_{j=1, j\neq i}^{n} x_{j}^{p_j}f({\bf X})d{\bf X}\vspace{0.3cm}\\
        \quad+\hspace{-0.1cm}\int_{\mathbb{R}^n}\prod_{j=1}^{n} x_{j}^{p_j}\dot{f}({\bf X})d{\bf X}
        +\hspace{-0.1cm}\int_{\mathbb{R}^n}\prod_{j=1}^{n} x_{j}^{p_j}f({\bf X})\left(\sum_{i=1}^{n}\frac{\partial \dot{x}_i}{\partial x_i}\right) d{\bf X}
  \end{array}
\end{equation}
Because the matrix ${\bf L}$ is antisymmetric (the diagonal entries are null), $ \frac{\partial \dot{x}_i}{\partial x_i}=0$,  the third term of eq.~(\ref{dmp1ppn}) vanishes when the distribution is subjected to rotational speeds. 
If we assume that the time derivative of the density function $\dot{f}(\mathbf{x})=0$, i.e. the density function in an infinitesimal point  does not change under rotation motions, the second term vanishes as well.
According to eq.~(\ref{Rotationsvar}), we have $\dot{x}_i=\sum_{j=1}^{n}l_{ij}x_j$, where $l_{ij}$ are the entries of the matrix ${\bf L}$ and combining with eq.~(\ref{dmp1ppn}), the time variation of moment caused by rotational speeds is obtained as,
\begin{equation}
	\label{dmp1ppnsimp}
   \begin{array}{ll}
		\dot{m}_{p_1 \cdots  p_n}&=\sum_{i=1}^{n}\sum_{j=1}^{n} p_i l_{ij} \int_{\mathbb{R}^n} x_i^{p_{i}-1} x_j \prod_{k=1, k\neq i}^{n} x_{k}^{p_k}f({\bf X})d{\bf X} \\
		&=\sum_{i=1}^{n}\sum_{j=1}^{n}p_il_{ij}{m}_{p_1,\cdots,p_i-1,\cdots,p_j+1,\cdots,p_n}.
	  \end{array}
\end{equation}
Using eq.~(\ref{dmp1ppnsimp}) and eq.~(\ref{A3D}) leads in 3D space to,
\begin{equation}
	\label{dmp1ppnsimp3D}
    \begin{array}{l}
		\dot{m}_{p_1p_2p_3}=\left(p_3m_{p_1,p_2+1,p_3-1}-p_2m_{p_1,p_2-1,p_3+1}\right)\omega_1\\
		\quad+\left(p_1m_{p_1-1p_2p_3+1}-p_3m_{p_1+1p_2p_3-1}\right)\omega_2\\
        \quad+\left(p_2m_{p_1+1p_2-1p_3}-p_1m_{p_1-1p_2+1p_3}\right)\omega_3
	\end{array},
\end{equation}

\noindent the time variational moment $\dot{m}_{p_1p_2p_3}$.

\section{\emph{N}-tuple descriptors definition}
Let assume the existence of a manifold with equivariance orthogonal transformation properties with respect to $\mathbf{X}$. 
The elements ${\bf x}_{\bf{v}}$ of this manifold, called \emph{n}-tuples, will be described as the linear combinations between a moments vector ${\bf v}^{(kk'\cdots)}_{(pp'\cdots)}$ and a set of parameters vectors $\boldsymbol{\alpha}_i$, such as ${\bf{x}_v}=\left[{{\boldsymbol{\alpha}}_1} {\bf v}^{(kk'\cdots)}_{(pp'\cdots)} \quad {\boldsymbol{\alpha}}_2 {\bf v}^{(kk'\cdots)}_{(pp'\cdots)} \quad\cdots \quad \boldsymbol{\alpha}_n {\bf v}^{(kk'\cdots)}_{(pp'\cdots)}\right]^{\top}$.

If ${\bf v}^1_p$ is the vector composed by all moments $m_{p_1 \cdots p_n}$ of order $p$ (described in section 2), that is, only monomials (from the linear combination between different moments) of degree 1. 
${\bf v}^2_p$ is defined as the  vector composed of the monomials of degree 2 resulting of the combination ${\bf v}^1_p\cdot{\bf v}^1_p$.
Follows straight-forward  to the $k$-times combination of ${\bf v}^1_p$ which obtains a vector ${\bf v}^k_p$ with monomials of degree $k$.
This can be extended to the resulting vector ${\bf v}^{(kk')}_{(pp')}$ by the combination of two different order-degree moments vectors ${\bf v}^k_p$ and ${\bf v}^{k'}_{p'}$.
Which generalizes to ${\bf v}^{(kk'\cdots)}_{(pp'\cdots)}$ for the combination of multiples moments vectors.

Because we assume equivariance orthogonal transformation condition, we can do a similar reasoning on the orthogonality condition of $n$-tuples using rotational speed.
Equivalent to eq.~(\ref{Rotationsvar}), the motion of a $n$-tuple ${\bf x_v}$ is formulated as,
\begin{equation}
	\label{eqspeed}
	\dot{ {\bf x}}_{\bf v}={\bf L}{\bf x_v}.
\end{equation}
\noindent Solving eq.~(\ref{eqspeed}) and assuming constant speed, i.e. constant matrix ${\bf L}$, we have the following relation,
\begin{equation}
	\label{Rotationder}
	{\bf x}_{\bf v}(t)=\exp\left({\bf L}_t\right){\bf v}_{\bf v}(0),
\end{equation}
\noindent where $\exp({\bf L}_t)$ is the matrix exponentiation of ${\bf L}_t$, ${\bf x}_{\bf v}(0)$ and ${\bf x}_{\bf v}(t)$ are respectively the initial and $t$ value of the $n$-tuple.
Knowing that $\exp({\bf L}_t)=\mathbf{R}$, eq.~(\ref{Rotationder}) can be rewritten as,
\begin{equation}
	\label{Rotationreal}
	{\bf x_v}(t)={\bf R}{\bf x_v}(0).
\end{equation}

\section{Orthogonal transformation estimation using \emph{n}-tuples}
Using the above properties
Details of the section, say that for the lack of space we will focuss on 3D case,
Because the lack of space we will focus the derivation on 3D space \emph{n}-tuples, called triplet.
Thus, triples are formally described as ${{{\bf x}}_{{\bf{v}}}}=[\boldsymbol{\alpha}_1^\top{\bf{v}}\quad\boldsymbol{\alpha}_2^\top{\bf{v}}\quad\boldsymbol{\alpha}_3^\top{\bf{v}}]^\top\in\mathbb{R}^3$.

\subsection{Orthogonal transformations and n-tuples}
Let us consider the case of 3D distributions, defined by ${\bf X}\subset\mathbb{R}^3$.
The central moments of a distribution eq.~(\ref{MomentsNDdiscrete}) after rotational motion, eq.~(\ref{Rotations}), can be expressed as a linear combination between a set of original moments and a random orthogonal transformation, eq.~(\ref{Rotationreal}). 
For example, taken the simplest triplet ${\bf x_v}=[m_{100}\quad m_{010}\quad m_{001}]^{\top}$ where $\mathbf{v}={\bf v}^{(1)}_{(1)}$, we can write that,
\begin{eqnarray}
    \label{exampletriplet2}
   \left[
        \begin{array}{c}
            m_{100}' \\
            m_{010}' \\
            m_{001}'
        \end{array}
        \right]
   =\left[
        \begin{array}{c}
            r_{11}m_{100}+r_{12}m_{010}+r_{13}m_{001} \\
            r_{21}m_{100}+r_{22}m_{010}+r_{23}m_{001} \\
            r_{31}m_{100}+r_{32}m_{010}+r_{33}m_{001}
        \end{array}
        \right]=
        {\bf R}{\bf x_v},
\end{eqnarray}
\noindent where $r_{ij}$ refers to the element of column $j$ and row $i$ of the matrix $\mathbf{R}$.
Unfortunately, eq.~(\ref{exampletriplet2}) is null for all possible ${\bf R}$ since central moments of order $p=1$ are null.
To overcome this problem, the chosen moments of ${\bf x_v}$ must be of order $p > 1$.
As discussed above, it is possible to define different triplets from moments of higher order.
For example, a consistent triplet with our hypothesis is ${\bf x_v}=\left[m_{300}+m_{120}+m_{102}\quad m_{210}+m_{030}+m_{012}\quad m_{201}+m_{021}+m_{003}\right]^{\top}$ which uses $\mathbf{v}={\bf v}^{(1)}_{(3)}$.
According to eq.~(\ref{MomentsND}), the first term of ${\bf x_v}$ is defined as,
\begin{equation}
    \label{exampletriplet1}
    m_{300}+m_{120}+m_{102}=\int_{-\infty}^{\infty}\int_{-\infty}^{\infty}\int_{-\infty}^{\infty} x_1\left(x^{2}_1+x^{2}_2+x^{2}_3\right)~d{\bf X}.
\end{equation}
From eq.~(\ref{exampletriplet1}) arise the relation between the first component of $\mathbf{X}$, i.e. $x_1$, and the first component of its triplet ${\bf x_v}$, i.e. $m_{300}+m_{120}+m_{102}$. Applying eq.~(\ref{exampletriplet1}) to eq.~(\ref{exampletriplet2}) we obtain that,
\begin{equation}
\label{exampletriplet11}
    \begin{array}{l}
   m_{300}'+m_{120}'+m_{102}'  = \int_{-\infty}^{\infty }\int_{-\infty}^{\infty }\int_{-\infty}^{\infty } x'_1\left({x{'}}^{2}_1+{x{'}}^{2}_2+{x{'}}^{2}_3\right)d{\bf X} \\
   =\int_{-\infty}^{\infty }\int_{-\infty}^{\infty }\int_{-\infty}^{\infty } \left(r_{11}x_1+r_{12}x_2+r_{13}x_3\right)\left(x^{2}_1+x^{2}_2+x^{2}_3\right)d{\bf X}\\
   =r_{11}\left(m_{300}+m_{120}+m_{102}\right) + r_{12}\left(m_{210}+m_{030}+m_{012}\right) \\
  \quad + r_{13}\left(m_{201}+m_{021}+m_{003}\right)
    \end{array}
\end{equation}
\noindent for the first component.
The same approach is applied to the successive components of ${\bf x_v}$ as, 
\begin{equation}
	\label{exampletriplet112}
	\left\{\begin{array}{l}
        m_{210}'+m_{030}'+m_{012}'=r_{21}(m_{300}+m_{120}+m_{102}) \\
        \quad +r_{22}(m_{210}+m_{030}+m_{012})  +r_{23}(m_{201}+m_{021}+m_{003}) \\
		 m_{201}'+m_{021}'+m_{003}'=r_{31}(m_{300}+m_{120}+m_{102}) \\
        \quad +r_{32}(m_{210}+m_{030} +m_{012})+r_{33}(m_{201}+m_{021}+m_{003}) \\
   \end{array}\right.
   \end{equation}
\noindent leading us to the proof  of equivariance condition for the rest of components.
Now, let us in the next following provide a general scheme for defining the n-tuples. 

\subsection{N-tuple derivation: 3D space case} 
Defined above the moment time variation, let us consider the following vectors of moments,
\begin{eqnarray*}
		{{\bf v}_2^1}&=&\left[m_{200}\quad  m_{110} \quad  \cdots \quad  m_{002}\right]^\top\\
		{{\bf v}_3^1}&=&\left[m_{300}\quad  m_{210}\quad  m_{201}\quad  \cdots\quad  m_{003}\right]^\top\\
		{{\bf v}_{(2,3)}^{(1,1)}}&=&\left[m_{300}m_{200}\quad  m_{300}m_{110}\quad  \cdots \quad  m_{003}m_{002}\right]^\top
\end{eqnarray*}
\noindent where ${\bf v}_2^1 \in \mathbb{R}^6$ is composed by the $6$ moments of order $2$, ${\bf v}_3^1 \in \mathbb{R}^{10}$ is composed by the $10$ moments of order $3$ and ${\bf v}_{(2,3)}^{(1,1)} \in \mathbb{R}^{60}$ is composed by the $60$ monomials products between the entries of ${\bf v}_2^1$ and ${\bf v}_3^1$.
By examining the monomial component $m_{030}m_{200}$ of vector ${\bf v}_{(2,3)}^{(1,1)}$ using eq.~(\ref{dmp1ppnsimp3D}), we have that,
\begin{equation}
\label{exampletripletcalculation1}
 \begin{array}{l}
 		\frac{d(m_{030}m_{200})}{dt}=m_{200}\dot{m}_{030}+m_{030}\dot{m}_{200}=-3m_{200}m_{021}\omega_1\vspace{0.2cm}\\
        \quad+2m_{101}m_{030}\omega_2 +(3m_{120}m_{200}-2m_{110}m_{030})\omega_3.
 \end{array}
 \end{equation}
Note that the coefficients of $\omega_i$ are linear combinations between moments of orders $2$ and $3$.
This keeps true for the time derivative of all entries of ${\bf v}={\bf v}_{(2,3)}^{(1,1)}$.
What allows us to express in a general form the time derivative of $\bf{v}$, as,
\begin{equation}
	\label{exampletripletcalculation2}
	\dot{{\bf{v}}}=({\bf L}^{\omega_1}_{\bf{v}}{\bf{v}})\omega_1+({\bf L}^{\omega_2}_{\bf{v}}\bf{v})\omega_2+({\bf L}^{\omega_3}_{\bf{v}}\bf{v})\omega_3,
\end{equation}

\noindent where ${\bf L}^{\omega_1}_{\bf{v}}$, ${\bf L}^{\omega_2}_{\bf{v}}$ and ${\bf L}^{\omega_3}_{\bf{v}}$ are matrices in $\mathbb{R}^{n\times n}$ space, where $n$ is the number of monomials in the vector ${\bf v}\in\mathbb{R}^{n}$, e.g. ${\bf L}^{\omega_1}_{\bf{v}} \in \mathbb{R}^{60\times60}$ for the case of ${\bf v}={\bf v}_{(2,3)}^{(1,1)}$.
These matrices represent the interaction between time variational rotational motion $\left[\omega_1\quad\omega_2\quad\omega_3\right]$ and the entries of ${\bf v}$, as was the case for $m_{030}m_{200}$.
Therefore, the time variation of ${\bf x}_{\bf v}=\left[ \boldsymbol{\alpha}_1^\top{\bf v}\quad  \boldsymbol{\alpha}_2^\top{\bf v}\quad  \boldsymbol{\alpha}_3^\top{\bf v}\right]^\top$ is given by,
\begin{equation}
	\label{exampletriplecalculation3}
	\dot{\bf x}_{\bf v}=\left[
		\begin{array}{ccc}
			\boldsymbol{\alpha}_1^\top{\bf L}_{\bf{v}}^{\omega_1}{\bf{v}}  & \boldsymbol{\alpha}_1^\top{\bf L}_{\bf{v}}^{\omega_2}{\bf{v}} & \boldsymbol{\alpha}_1^\top{\bf L}_{\bf{v}}^{\omega_3}{\bf{v}} \\
			\boldsymbol{\alpha}_2^\top{\bf L}_{\bf{v}}^{\omega_1}{\bf{v}}  & \boldsymbol{\alpha}_2^\top{\bf L}_{\bf{v}}^{\omega_2}{\bf{v}} & \boldsymbol{\alpha}_2^\top{\bf L}_{\bf{v}}^{\omega_3}{\bf{v}} \\
			\boldsymbol{\alpha}_3^\top {\bf L}_{\bf{v}}^{\omega_1}{\bf{v}} & \boldsymbol{\alpha}_3^\top{\bf L}_{\bf{v}}^{\omega_2}{\bf{v}} & \boldsymbol{\alpha}_3^\top{\bf L}_{\bf{v}}^{\omega_3}{\bf{v}}
		\end{array}
		\right]
	\left[
		\begin{array}{c}
			\omega_1 \\
			\omega_2 \\
			\omega_3
		\end{array}
		\right]
\end{equation}
\noindent Then, we have $9$ conditions if eq.~(\ref{exampletriplecalculation3}) is of the same form as eq.~(\ref{A3D}):
\begin{eqnarray*}
	\label{exampletriplecalculation4}
    \hspace{-0.8cm}
	\begin{array}{lll}
			 1)~\boldsymbol{\alpha}_1^\top{\bf L}_{\bf{v}}^{\omega_1}{\bf{v}}=0
			 & 2)~\boldsymbol{\alpha}_2^\top{\bf L}_{\bf{v}}^{\omega_1}{\bf{v}}=\boldsymbol{\alpha}_3^\top{\bf{v}}
			 & 3)~\boldsymbol{\alpha}_3^\top{\bf L}_{\bf{v}}^{\omega_1}{\bf{v}}=-\boldsymbol{\alpha}_2^\top{\bf{v}} \\
			 4)~\boldsymbol{\alpha}_1^\top{\bf L}_{\bf{v}}^{\omega_2}{\bf{v}}=-\boldsymbol{\alpha}_3^\top{\bf{v}} 
			 & 5)~\boldsymbol{\alpha}_2^\top{\bf L}_{\bf{v}}^{\omega_2}{\bf{v}}=0
			 & 6)~\boldsymbol{\alpha}_3^\top{\bf L}_{\bf{v}}^{\omega_2}{\bf{v}}=\boldsymbol{\alpha}_1^\top{\bf{v}}  \\
			 7)~\boldsymbol{\alpha}_1^\top{\bf L}_{\bf{v}}^{\omega_3}\bf{v}=\boldsymbol{\alpha}_2^\top{\bf{v}}    
			 & 8)~\boldsymbol{\alpha}_2^\top{\bf L}_{\bf{v}}^{\omega_3}{\bf{v}}=-\boldsymbol{\alpha}_1^\top{\bf{v}} 
			 & 9)~\boldsymbol{\alpha}_3^\top{\bf L}_{\bf{v}}^{\omega_3}{\bf{v}}=0
	\end{array}
\end{eqnarray*}
\noindent The constraints  must be valid for any value of ${\bf v}$. 
They can be written under the form,
\begin{equation}
	\label{exampletriplecalculation6}
	\left[
		\begin{array}{ccc}
			{\bf L}^{\omega_1\top}_{\bf{v} }    & {\bf 0}    & {\bf 0} \\
			{\bf L}^{\omega_1\top}_{\bf{v}}     & {\bf 0}    & -\mathbb{I} \\
			{\bf L}^{\omega_1\top}_{\bf{v}} & \mathbb{I} & {\bf 0} \\
			\vdots                              & \vdots     & \vdots \\
			{\bf 0}                             & {\bf 0}    & {\bf L}^{\omega_3\top}_{\bf{v}} \\
		\end{array}
		\right]
	\left[
		\begin{array}{ccc}
			\boldsymbol{\alpha}_1 \\
			\boldsymbol{\alpha}_2 \\
			\boldsymbol{\alpha}_3
		\end{array}
		\right]={\bf 0},
\end{equation}

\noindent where $\mathbb{I}$ and ${\bf 0}$ are respectively the identity and zero matrices of size $n\times n$ (e.g. ${\bf v}={\bf v}_{(2,3)}^{(1,1)}$ is $60\times60$). The overview of the complete procedure to derive $n$-tuples in 3D space is shown in Table~\ref{tab:n_tuples_computation}.

\begin{table}
    \centering
    \begin{tabular}{|p{8.4cm}|}
        \hline\vspace{-0.5cm}\begin{enumerate}
        \item Define the moment vector ${\bf v}={\bf v}^{(kk'k''...)}_{(pp'p"...)}$ such that its degree $p\cdot k+p'\cdot k'+...$ is an odd integer since even degrees resolve as null.
        \item Compute ${\bf L}_{\bf{v}}^{\omega_1}$, ${\bf L}_{\bf{v}}^{\omega_2}$ and ${\bf L}_{\bf{v}}^{\omega_3}$ using eq.~(\ref{dmp1ppnsimp3D}),.
        \item Solve eq.~(\ref{exampletriplecalculation6}) to obtain the vectors $\boldsymbol{\alpha}_1$, $\boldsymbol{\alpha}_2$ and $\boldsymbol{\alpha}_3$, which determine the number of $n$-tuples.
\end{enumerate}\nointerlineskip\\
\hline
    \end{tabular}
    \caption{Summarized procedure used to derive $n$-tuples.}
    \label{tab:n_tuples_computation}
\end{table}

In the case of 2D distribution, the same framework allows obtaining $n$-tuples ${\bf x}_{\bf v}=\left[\boldsymbol{\alpha}_1^\top {\bf v }\quad \boldsymbol{\alpha}_2^\top{\bf v }\right]^\top$, called duoblet, that behave as a point in a plane with respect to rotations.
As examples, using ${\bf v_{(2,3)}^{(1,1)}}$ in 3D space,  3 triplets can be generated:
\begin{small}
\begin{equation}
\label{P_3D}
\begin{array}{l}
\hspace{-0.5cm}
{\bf x_{v1}}=\left[\begin{array}{l}
        m_{003}m_{101}+ m_{012}m_{110}+ m_{021}m_{101}+ m_{030}m_{110}+ m_{101}m_{201}\\ 
        \quad+m_{102}m_{200}+ m_{110}m_{210}+ m_{120}m_{200}+ m_{200}m_{300} \\
        m_{003}m_{011}+ m_{011}m_{021}+ m_{012}m_{020}+ m_{020}m_{030}+ m_{011}m_{201}\\
        \quad+m_{102}m_{110}+ m_{020}m_{210}+ m_{110}m_{120}+ m_{110}m_{300} \\
        m_{002}m_{003}+ m_{002}m_{021}+ m_{011}m_{012}+ m_{011}m_{030}+ m_{002}m_{201}\\
        \quad+m_{101}m_{102}+ m_{011}m_{210}+ m_{101}m_{120}+ m_{101}m_{300}
    \end{array}\right]\\
   \hspace{-0.5cm} {\bf x_{v2}}=\left[\begin{array}{l}
            m_{002}m_{120}- 2m_{011}m_{111}+ m_{020}m_{102}+ m_{002}m_{300}- 2m_{101}m_{201}\\
            +m_{102}m_{200}+ m_{020}m_{300}- 2m_{110}m_{210}+ m_{120}m_{200} \\
            m_{002}m_{030}- 2m_{011}m_{021}+ m_{012}m_{020}+ m_{002}m_{210}+ m_{012}m_{200}\\
            \quad-2m_{101}m_{111}+ m_{020}m_{210}+ m_{030}m_{200}- 2m_{110}m_{120} \\
            m_{002}m_{021}+ m_{003}m_{020}- 2m_{011}m_{012}+ m_{002}m_{201}+ m_{003}m_{200}\\
            \quad-2m_{101}m_{102}+ m_{020}m_{201}+ m_{021}m_{200}- 2m_{110}m_{111}
        \end{array}\right]\\
    {\bf x_{v3}}=\left[\begin{array}{l}
            m_{002}m_{102 }+ 2m_{011}m_{111 }+ m_{020}m_{120 }+ 2m_{101}m_{201}\\
            \quad+2m_{110}m_{210 }+ m_{200}m_{300} \\
            m_{002}m_{012 }+ 2m_{011}m_{021 }+ m_{020}m_{030 }+ 2m_{101}m_{111}\\
            \quad+2m_{110}m_{120 }+ m_{200}m_{210} \\
            m_{002}m_{003 }+ 2m_{011}m_{012 }+ m_{020}m_{021 }+ 2m_{101}m_{102}\\
            \quad+2m_{110}m_{111 }+ m_{200}m_{201}
        \end{array}\right]
    \end{array}
\end{equation}
\end{small}

In 2D space, using the vector of moments ${\bf v_{(2,3)}^{(1,1)}}$, 3 doublet can be generated as: 
\begin{small}
\begin{equation}
\label{P2D23v1}
    \begin{array}{l}
        {\bf x_{v_1}}=\left[
        \begin{array}{c}
            m_{03}m_{11 }- m_{02}m_{12 }- m_{02}m_{30 }+ m_{11}m_{21} \\
            m_{11}m_{12 }- m_{03}m_{20 }+ m_{11}m_{30 }- m_{20}m_{21}
        \end{array} \right] \\
        {\bf x_{v_2}}=\left[
        \begin{array}{c}
            2m_{03}m_{11 }-2m_{02}m_{12 }- m_{02}m_{30 }+ m_{12}m_{20} \\
            m_{02}m_{21 }- m_{03}m_{20 }+2m_{11}m_{30 }-2m_{20}m_{21}
        \end{array}\right]\\
        {\bf x_{v_3}}=\left[
        \begin{array}{c}
            3m_{02}m_{12 }-2m_{03}m_{11 }+2m_{02}m_{30}+ m_{20}m_{30} \\
            m_{02}m_{03 }+2m_{03}m_{20}-2m_{11}m_{30 }+ 3m_{20}m_{21}
        \end{array} \right]
    \end{array}
\end{equation}
\end{small}

An extended example is given in the script\footnote{\url{https://drive.google.com/file/d/1UF-5FANGIxAnVq1jLQm2h_f8d3BxpRo7/view?usp=sharing}} which provides a list of n-tuples computed from moments of different orders and in different original spaces 2-, 3- and 4-dimension.

\subsection{Reflection transformation and symmetry}
A reflection transformation is defined by a sign flip of one coordinate while leaving the rest unchanged.
An example of reflection is the transformation defined by ${\bf x}'={\bf F}{\bf x}$ subject to the coordinates change $x_1'=x_1$,  $x_2'=-x_2$ and $x_3'=x_3$.
As for rotations, a reflection ${\bf F}$ holds ${\bf F}^\top {\bf F}={\bf I}$, but with det$({\bf F})=-1$.
The effect of such transformation on a moment is given by,
\begin{equation}
	\label{momentreflection}
	m_{p_1p_2p_3}'=(-1)^{p_2}m_{p_1p_2p_3}.
\end{equation}

\noindent If the sign changing was on $x_1$, the effect on moment would be $m_{p_1p_2p_3}'=(-1)^{p_1}m_{p_1p_2p_3}$.
If a $n$-tuple undergoes the same reflection as the object points, the vectors of coefficients will have to hold $
	\boldsymbol{\alpha}_1 {\bf v}'=\boldsymbol{\alpha}_1 {\bf v},  
	\boldsymbol{\alpha}_2 {\bf v}'=-\boldsymbol{\alpha}_2 {\bf v}$ and $
	\boldsymbol{\alpha}_3 {\bf v}'=\boldsymbol{\alpha}_3 {\bf v}
$, where ${\bf v}'$ is obtained from ${\bf v}$ using eq.~(\ref{momentreflection}).
Surprisingly, for the different set of $n$-tuples obtained by solving eq.~(\ref{exampletriplecalculation6}), they already hold the reflection conditions.

Now, lets us assume that the object is symmetrical with respect to the plane $x_1x_3$.
For such object, the considered reflection example will leave its shape unchanged.
Since the moment $n$-tuples undergo the same reflection, they must belong to the plane $x_1x_3$.
Otherwise, one can make difference between the distribution configurations before and after reflection.
In fact, this is valid for any plane of symmetry, since the $n$-tuples undergo the same rotation as that applied to an object.
The same reasoning can be used for other kinds of symmetries.

Note that we can obtain the symmetry plane via Singular Value Decomposition (SVD) of the $n$-tuple. Concretely, the singular vector that describes the symmetry plane is the one associated with minimal singular value.
The type of symmetry is determined by the relationship among the different singular values.
In addition, two non-aligned triplets are enough to estimate orthogonal transformation (rotation or reflection) since they can build an orthogonal base of the object.
But, with more than two triplets the rotations can be found using Procrustes methods~\cite{schonemann1966generalized}.

\section{Experiments in 2D space}
\begin{figure}[t]
	\centerline{
		\includegraphics[width=.25\columnwidth]{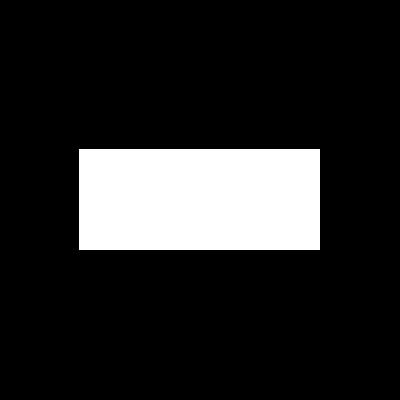}
  \hspace{0.2cm}
  \includegraphics[width=.25\columnwidth]{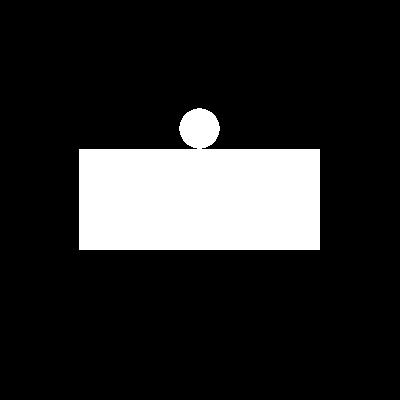}
  \hspace{0.2cm}
  \includegraphics[width=.25\columnwidth]{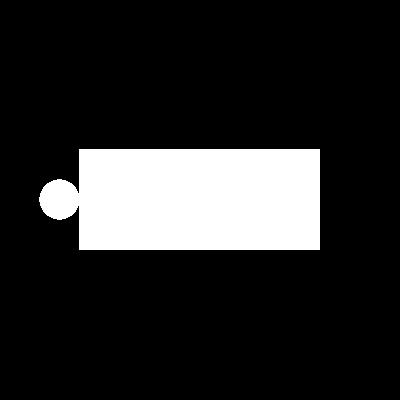}}
	\centerline{		
 \includegraphics[width=.3\columnwidth]{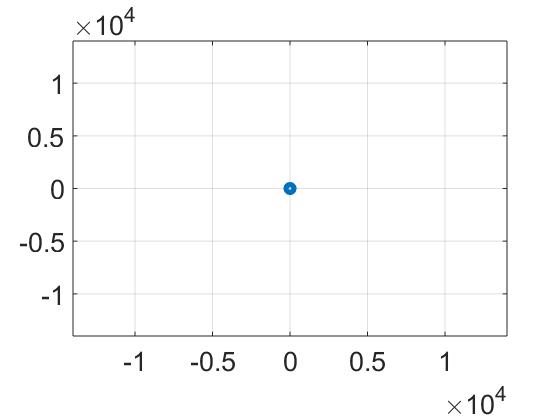}
		\includegraphics[width=.3\columnwidth]{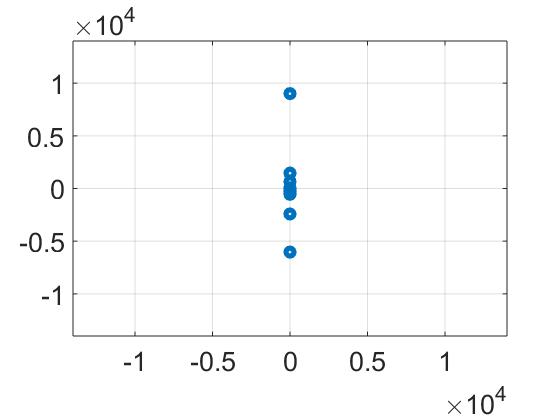}
		\includegraphics[width=.3\columnwidth]{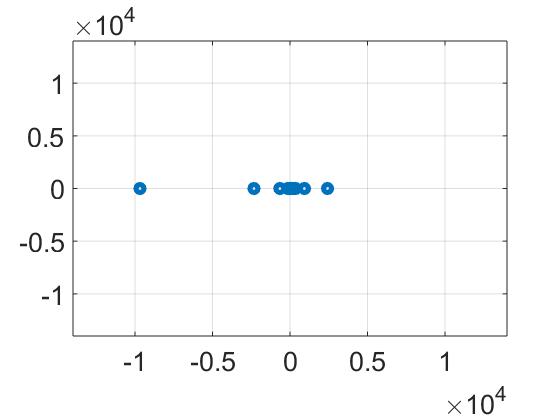}}
	\centerline{(a) \hspace{2cm} (b) \hspace{2cm} (b) }
	\caption{Case study of symmetry: (a) object center, (b) x-axis (c) y-axis}
	\label{casestudy}
\end{figure}

We examine how the $n$-tuples in 2D behave under the effect of adding non-symmetry transformation on a controlled example (Fig.~\ref{casestudy}).
For each image, doublets are computed from the moment vectors ${\bf v}_{(2,3)}^{(1,1)}$, ${\bf v}_{(2,5)}^{(1,1)}$, ${\bf v}_{(4,3)}^{(1,1)}$, ${\bf v}_{(4,5)}^{(1,1)}$, ${\bf v}_{(6,3)}^{(1,1)}$, ${\bf v}_{(6,5)}^{(1,1)}$, ${\bf v}_{(6,7)}^{(1,1)}$, ${\bf v}_{(6,9)}^{(1,1)}$, ${\bf v}_{(8,7)}^{(1,1)}$ and ${\bf v}_{(8,9)}^{(1,1)}$.
For the rectangle given in Fig.~\ref{casestudy}(a) top, the corresponding $n$-tuples coincide with the origin since the object is symmetrical with respect to its center (Fig.~\ref{casestudy}(a) bottom).
For the two objects shown in Fig.~\ref{casestudy}(b) and Fig.~\ref{casestudy}(c), they exhibit symmetries with respect to two different axis.
They were obtained by adding a disc to the rectangle in specific positions.
Their doublets are shown in the bottom row of Fig.~\ref{casestudy}(b) and Fig.~\ref{casestudy}(c).
The plots show that their $n$-tuples form lines in the directions of their respective symmetry axis.
\begin{figure}[b]
	\centerline{
		\includegraphics[width=.2\columnwidth]{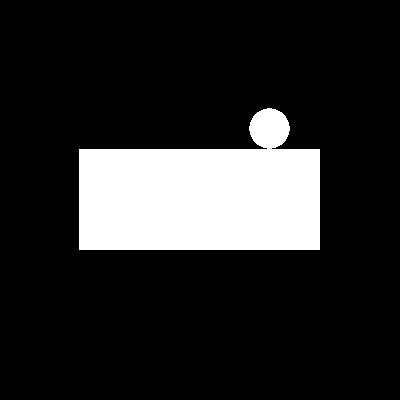}
		\includegraphics[width=.25\columnwidth]{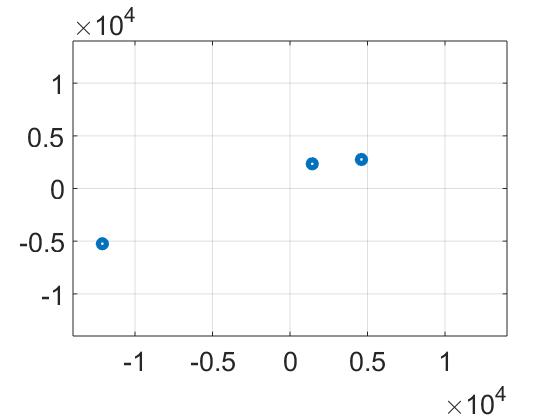}
		\includegraphics[width=.25\columnwidth]{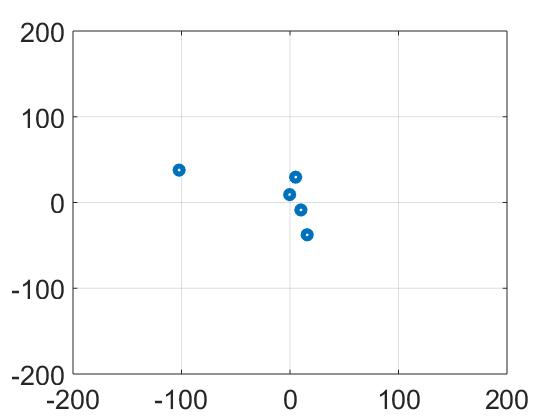}
		\includegraphics[width=.25\columnwidth]{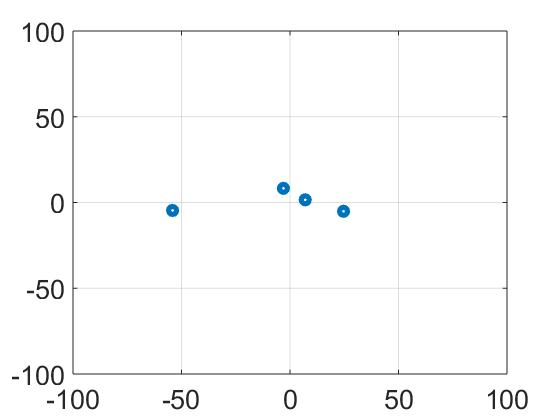}
	}
	\centerline{(a) \hspace{1.5cm} (b) \hspace{1.5cm} (c) \hspace{1.5cm} (d) }
	\caption{Case study of a non-symmetrical object: (a) the object, (b) $n$-tuples from ${\bf v}_{(2,3)}^{(1,1)}$,  (c) $n$-tuples from ${\bf v}_{(4,5)}^{(1,1)}$, (d) $n$-tuples from  ${\bf v}_{(6,3)}^{(1,1)}$ }
	\label{casestudy1}
\end{figure}

\begin{figure}[t]
	\centerline{
		\includegraphics[width=0.2\linewidth]{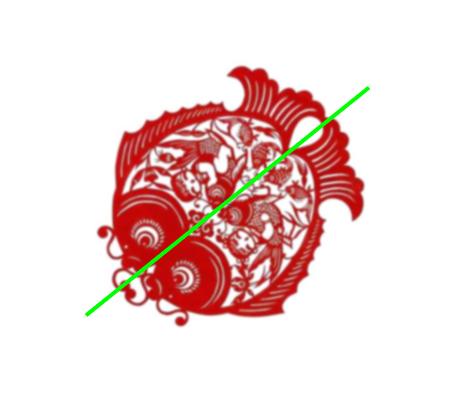}\hspace{-0.5cm}
		\includegraphics[width=0.2\linewidth]{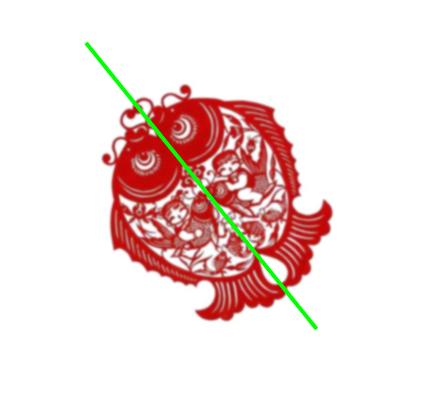}\hspace{-0.5cm}
		\includegraphics[width=0.2\linewidth]{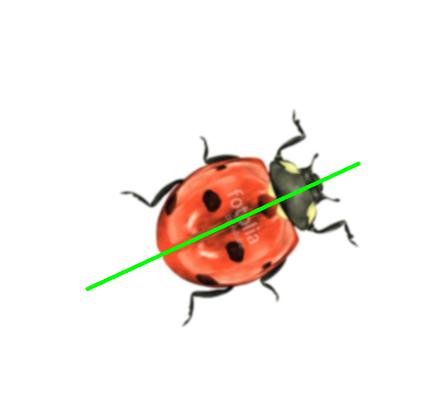}\hspace{-0.5cm}
		\includegraphics[width=0.2\linewidth]{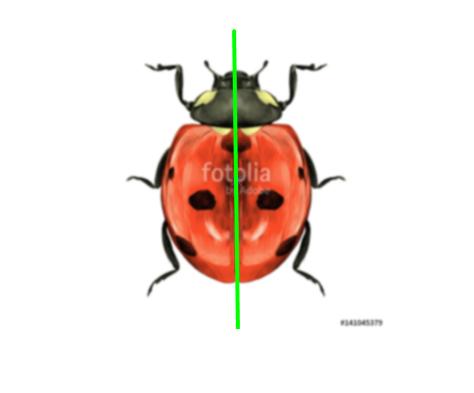}\hspace{-0.5cm}
		\includegraphics[width=0.2\linewidth]{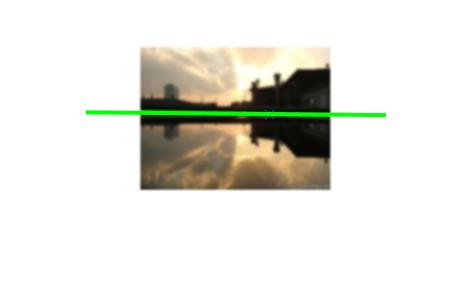}\hspace{-0.7cm}
		\includegraphics[width=0.2\linewidth]{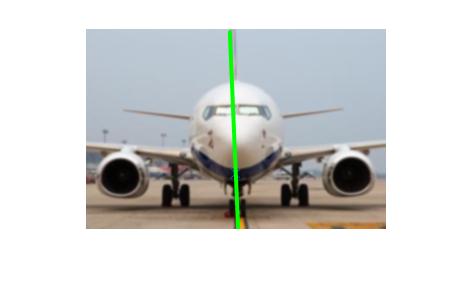}\hspace{-0.7cm}
		\includegraphics[width=0.2\linewidth]{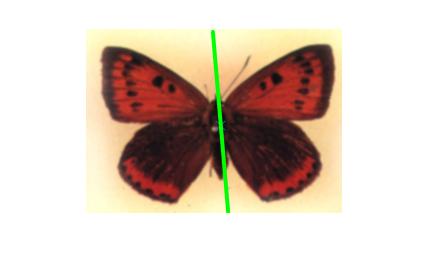}
	}
	\caption{Example of mirror symmetry}
	\label{examplemirror}
\end{figure}

In another experiment, shown in Fig.~\ref{casestudy1}(a), the considered object does not exhibit any symmetry.
This time, doublets were computed separately from the vector moments ${\bf v}_{(2,3)}^{(1,1)}$, ${\bf v}_{(4,5)}^{(1,1)}$, and ${\bf v}_{(6,3)}^{(1,1)}$.
The obtained results are shown in Fig.~\ref{casestudy1}(b) for ${\bf v}_{(2,3)}^{(1,1)}$,  Fig.~\ref{casestudy1}(c) for ${\bf v}_{(4,5)}^{(1,1)}$ and Fig.~\ref{casestudy1}(d) for ${\bf v}_{(6,3)}^{(1,1)}$.
The figures show that asymmetries are more noticeable for doublets computed from moment of higher order.
Indeed, the obtained ratios, using SVD decomposition, between the higher and smaller singular values of the doublets were $8.73$, $2.71$ and $5.59$ for ${\bf v}_{(2,3)}^{(1,1)}$, ${\bf v}_{(4,5)}^{(1,1)}$, and ${\bf v}_{(6,3)}^{(1,1)}$ respectively.
In general, the smaller a ratio is, farthest the $n$-tuples is for forming a line.
A last experiment, illustrated in Fig.~\ref{examplemirror}, explores approximate mirror symmetry.
For the considered objects, the images are first converted to gray-scale then to binary.
The doublets are computed using centered moments of the binary image.
Finally, the lines of symmetry are estimated using the direction corresponding to the smaller singular value of the doublets.

\section{Experiments in 3D space}
The goal of this part is to validate the symmetry detection as well as orthogonal transformations using triplets.
The used objects are part of the datasets McGill University \footnote{http://www.cim.mcgill.ca/~shape/benchMark/} and ShapeNet~\cite{chang2015shapenet}.
In both cases, objects are not completely symmetrical.
As for images case, $n$-tuples are computed from the central moments of different orders.  
\subsection{Symmetry with respect to a plane}
\begin{figure}[b]
	\centerline{
		\includegraphics[width=.3\columnwidth]{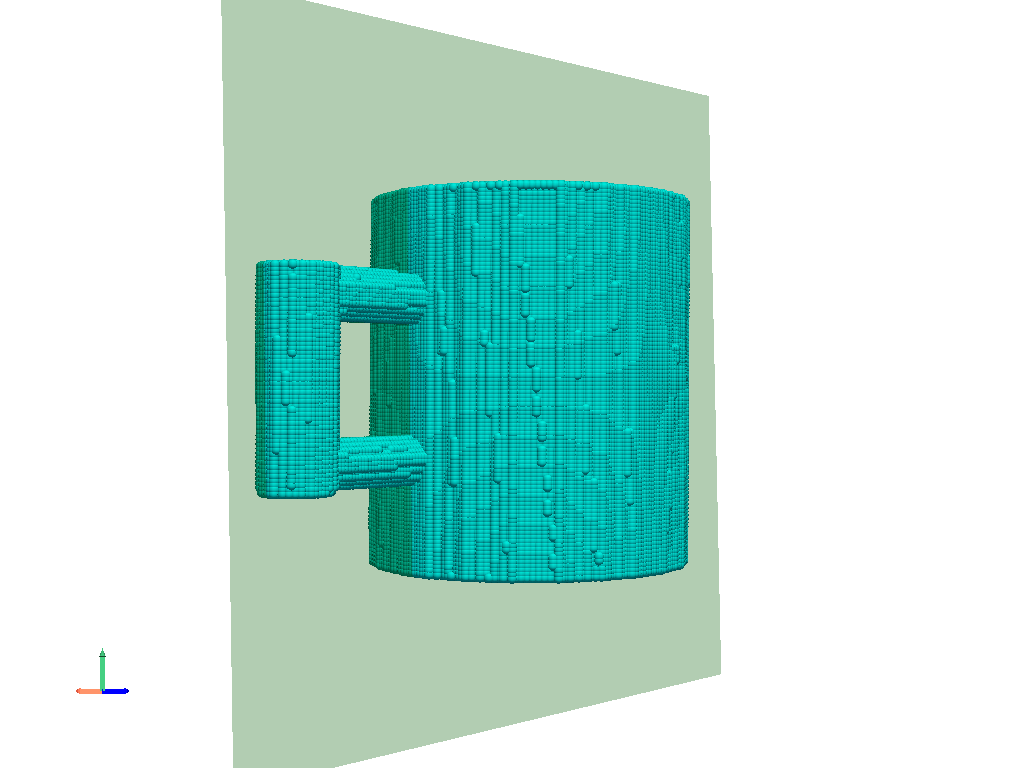}
        \hspace{0.2cm}
		\includegraphics[width=.3\columnwidth]{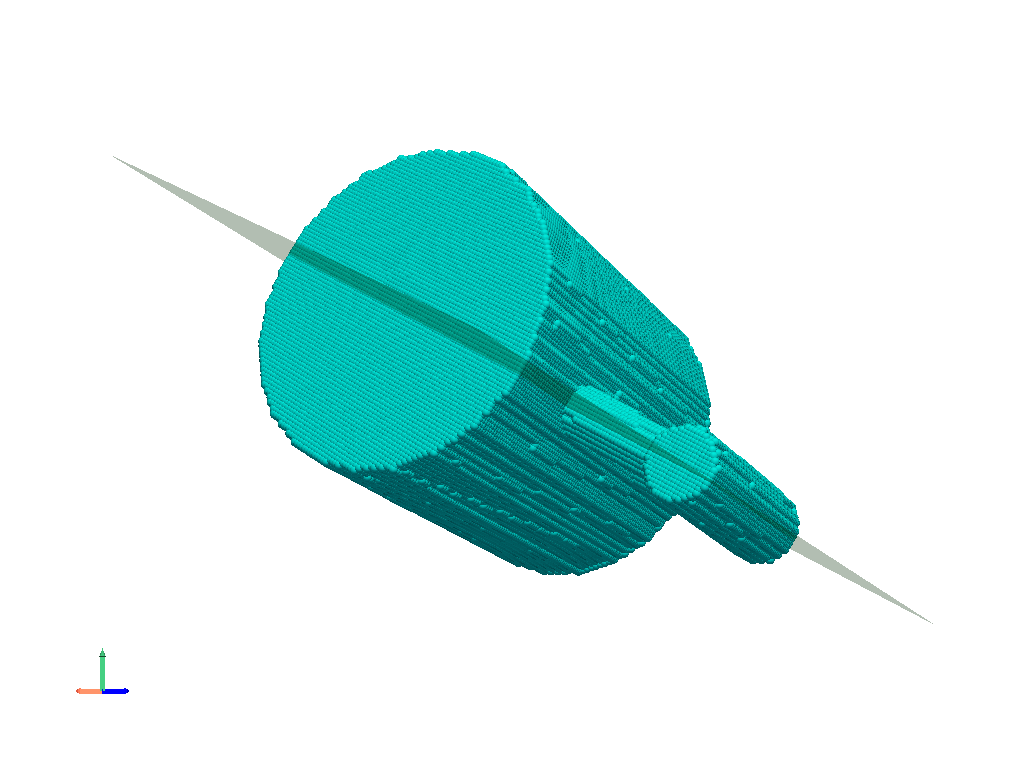}
        \hspace{0.2cm}
		\includegraphics[width=.3\columnwidth]{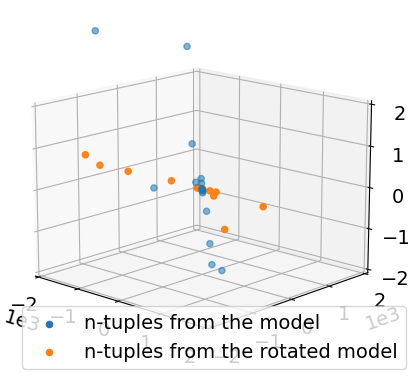}
  }
	\centerline{(a)\hspace{2.5cm}(b)\hspace{2.5cm} (c) }
	\caption{Mug, a symmetrical object case of study: (a) Plane of symmetry of the object with initial conditions, $\mathbf{R}=\mathbb{I}$. (b) Plane of symmetry recovered after being applied to a random $\mathbf{R}$ rigid transformation. (c) Representation of moment's $n$-tuples in Cartesian space.}
	\label{fig:cups2model}
\end{figure}

We deal with symmetry detection and rotation estimation in this experiment.
The chosen object to conduct this experiment is a random mug (Fig.~\ref{fig:cups2model}) from ShapeNet.
Fig.~\ref{fig:cups2model}(a) shows the object model in their original pose $\mathbb{I}\in \mathbb{R}^{3\times3}$ and Fig.~\ref{fig:cups2model}(b) shows it after being applied the rotation $\mathbf{R}\in$~SO(3) defined as follows,
\begin{equation}
	\label{Rcups2}
	{\bf R}=\left[
		\begin{array}{ccc}
			-0.3085 & 0.2118  & 0.9273  \\
			0.8599  & -0.3546 & 0.3671  \\
			0.4066  & 0.9106  & -0.0727
		\end{array}\right].
\end{equation}

For the two considered mug poses, the obtained singular vectors are respectively $\mathbf{V}$ and ${\mathbf{V_R}}$ (eq.~\ref{vsigulardecompositioncups2model}), holding two equal singular values $\mathbf{S} ={\mathbf{S_R}}$ (eq.~\ref{ssigulardecompositioncups2model}).
This shows the equivalent relationship between the geometric transformation of the model and the $n$-tuples transformation for the single plane symmetry case.
Since the shape of the distribution is hold during orthogonal transformation SO(3), that is, the singular values keep constants.
But their singular values are transformed equally to the geometric transformation of the 3D object model (Fig.~\ref{fig:cups2model}(c)).
\begin{equation}
	\label{vsigulardecompositioncups2model}
	{\bf V}=\left[
		\begin{array}{ccc}
			-0.001 & 0.027 & 0.999  \\
			-0.343 & 0.938 & -0.026 \\
			0.939  & 0.342 & -0.007
		\end{array}\right];
		\,\,\mathbf{V_R}=\bf{R\,V}
\end{equation}

\noindent In addition, it can be estimated the type of symmetry from the shape of the moment's $n$-tuples, i.e. singular values.
Let $\sigma_1$, $\sigma_2$, and $\sigma_3$ be the singular values, such as $\sigma_1 > \sigma_2 > \sigma_3$, and ${\bf{\nu}_1}$, ${{\bf\nu}_2}$, ${{\bf\nu}_3}$ their corresponding singular vectors.
The relation between the two smallest singular values and the biggest one determine the type of symmetry.
Such as, if the $\frac{\sigma_1}{\sigma_3}$ is orders of magnitude bigger than $\frac{\sigma_1}{\sigma_2}$, the symmetry is planar.
This is due to the existence of a dominant symmetry plane.
In the specific case, Fig.~\ref{fig:cups2model}(a), we have that $\frac{\sigma_1}{\sigma_3}=1121.22$ and $\frac{\sigma_1}{\sigma_2}=17.40$.
Meaning that, the $n$-tuples, triplets, form a plane of symmetry defined by ${\bf{\nu}_3}$, i.e. the third column of the matrix ${\bf V}$, eq.~(\ref{vsigulardecompositioncups2model}).
\begin{equation}
	\label{ssigulardecompositioncups2model}
	{\bf S_R}=\left[\begin{array}{ccc}
			3.2705 &
			0.1879 &
			0.0029
		\end{array}\right]\cdot 10^5 = {\bf S}
\end{equation}
\noindent Finally, the estimation of the orthogonal rotational transformation between both distributions, objects, poses is obtained as a generalized solution of the orthogonal Procrustes problem 
between the two set of moments $n$-tuples computed in Table~\ref{tab:n_tuples_computation}.
The real and estimated values of the rotation matrix are given by eq.~(\ref{Restcups2}), recovering the original transformation, eq.~(\ref{Rcups2}):
\begin{equation}
	\label{Restcups2}
	\widehat{\bf{R}}=\left[\begin{array}{ccc}
			-0.3085 & 0.2118  & 0.9273  \\
			0.8599  & -0.3546 & 0.3671  \\
			0.4066  & 0.9106  & -0.0727
		\end{array}\right] = {\bf{R}}
\end{equation}
\noindent The obtained result shows that the estimation is quite accurate.
Note also that ${{\bf V_R}}={\bf R} {\bf V}$.
Which means that the plane of symmetry undergoes the same rotation applied to the object.

\subsection{Symmetry with respect to a radial axis}
\begin{figure}[t]
	\centerline{
		\includegraphics[width=.3\columnwidth]{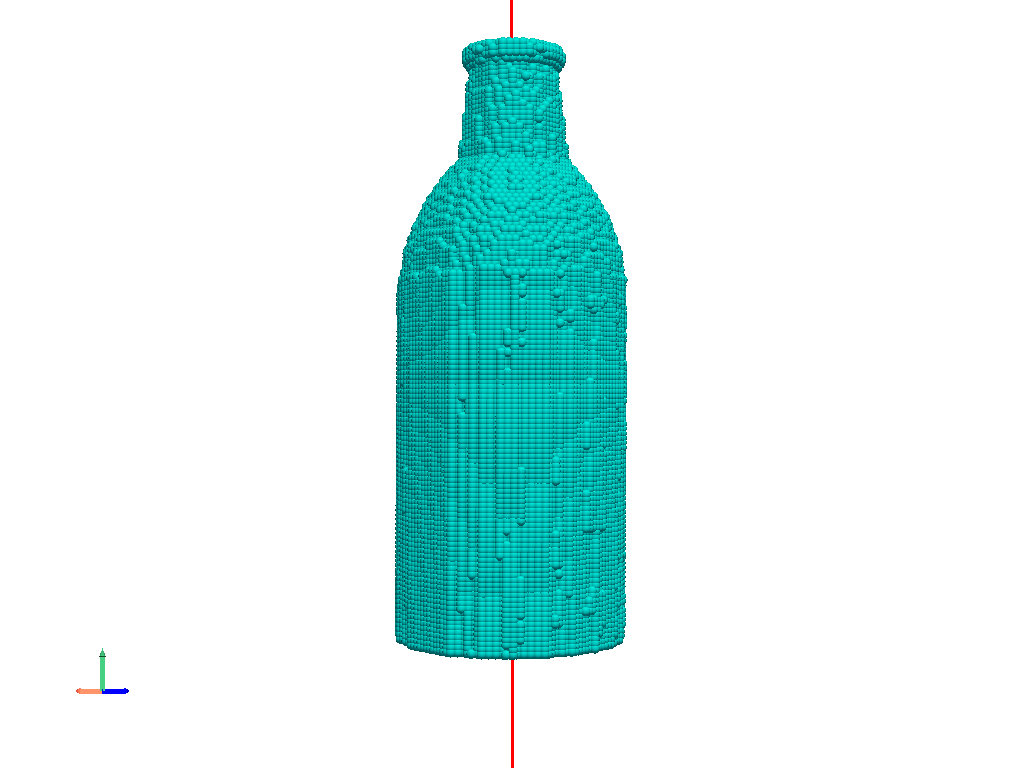}
        \hspace{-0.2cm}
		\includegraphics[width=.3\columnwidth]{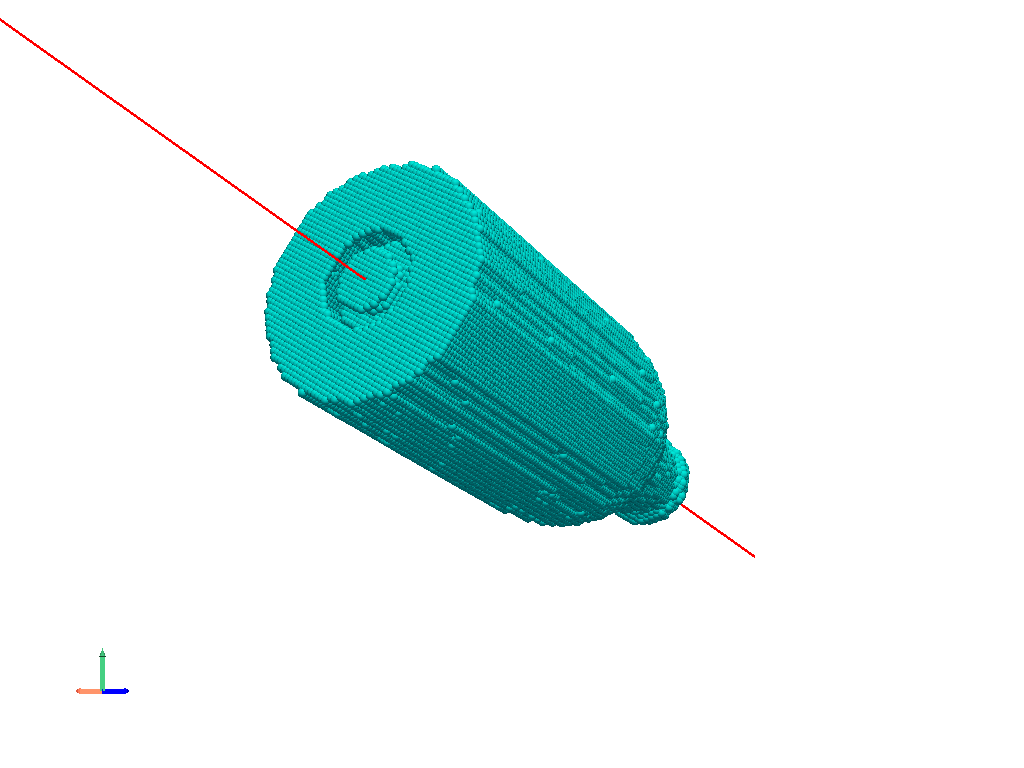}
        \hspace{-0.2cm}
		\includegraphics[width=.4\columnwidth]{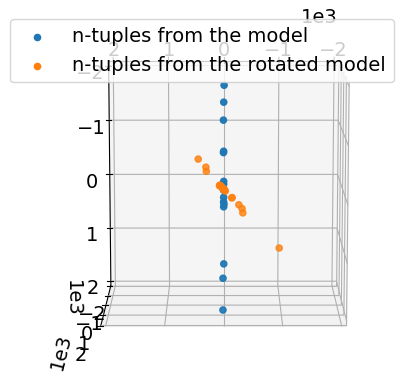}
	}
	\centerline{(a)\hspace{2.5cm}(b)\hspace{2.5cm} (c) }
	\caption{Axial symmetry:(a) bottle model, (b) the rotated bottle , (c) the $n$-tuples computed from moments }
	\label{fig:tableaxial1}
\end{figure}

In order to show how moments $n$-tuples deal with radial symmetry, we present the following case study.
Concretely, this experiment shows the model of a generic bottle with a radial symmetry axis (Fig.~\ref{fig:tableaxial1}(a)).
As for the previous case, both considered object poses are: 
aligned to the global frame, y-axis up-direction and z-axis forward direction; and obtained by applying a random transformation.

The ratios of singular values are $\frac{\sigma_1}{\sigma_{3}}=1926.33$ and $\frac{\sigma_{1}}{\sigma_{2}}=149.99$. Differently from the mug case (Fig.~\ref{fig:cups2model}), here the distance between both ratios is one order of magnitude smaller.
The intersection of the planes defined the two  singular vectors ${\bf{\nu}_2}$, ${\bf{\nu}_3}$ corresponding to the smaller singular values defines the symmetry axis as shown in Fig.~\ref{fig:tableaxial1}(a) and Fig.~\ref{fig:tableaxial1}(b). Finally, the $n$-tuples positions are shown in Fig.~\ref{fig:tableaxial1}(c).
They show that they almost belong to the symmetry axis for the two bottle poses.
\begin{equation}
	\label{ssigulardecompositiontable1model}
	{\bf S}=\left[\begin{array}{ccc}
			9.3904 & 0.0626 & 0.0048
		\end{array}\right]\cdot 10^4 = {\bf S_R}
\end{equation}

\subsection{Boosting b-plane symmetry by axial symmetry}
\begin{figure}[b]
	\centerline{
		\includegraphics[width=.3\columnwidth]{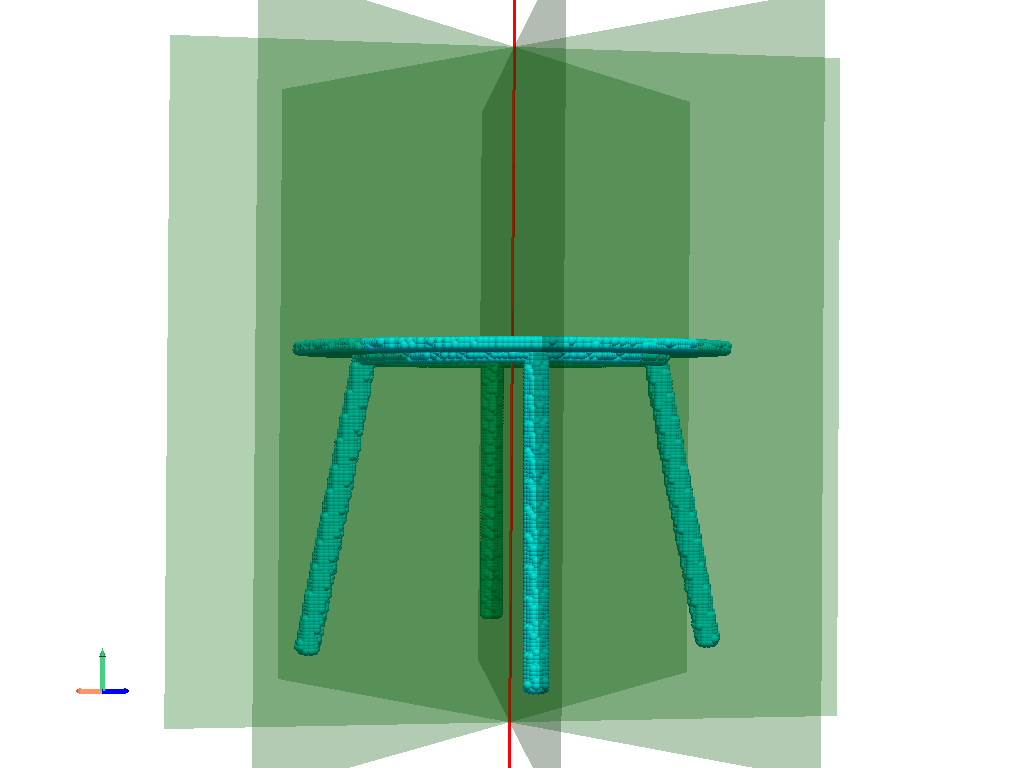}
		\includegraphics[width=.3\columnwidth]{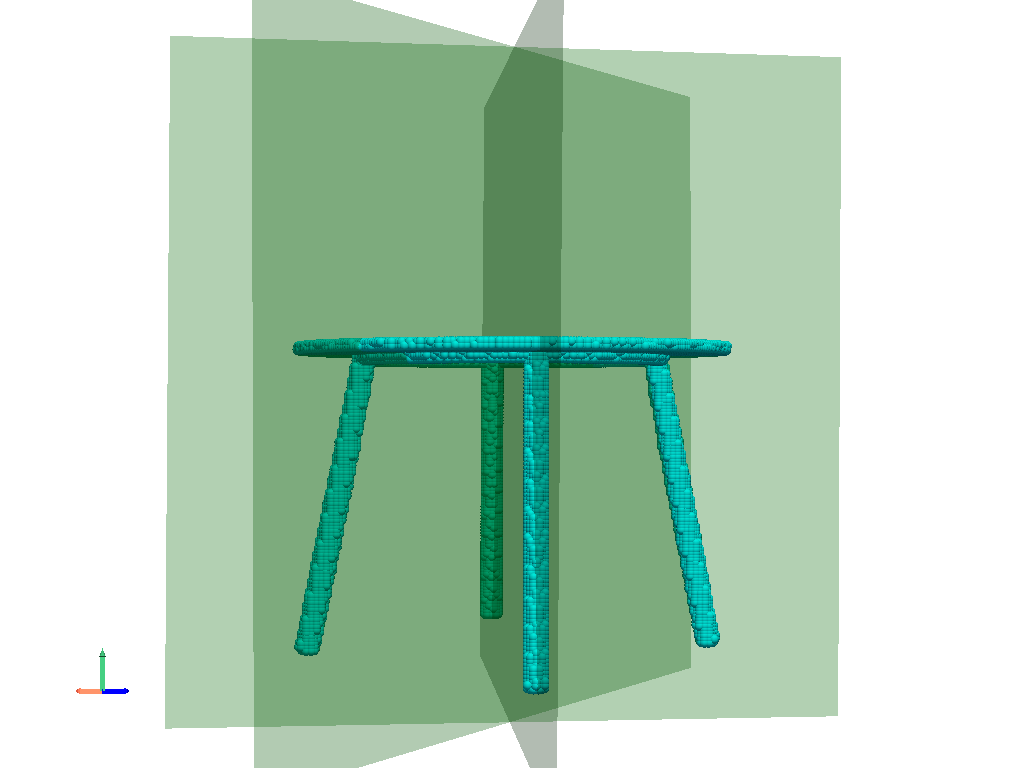}
		\includegraphics[width=.3\columnwidth]{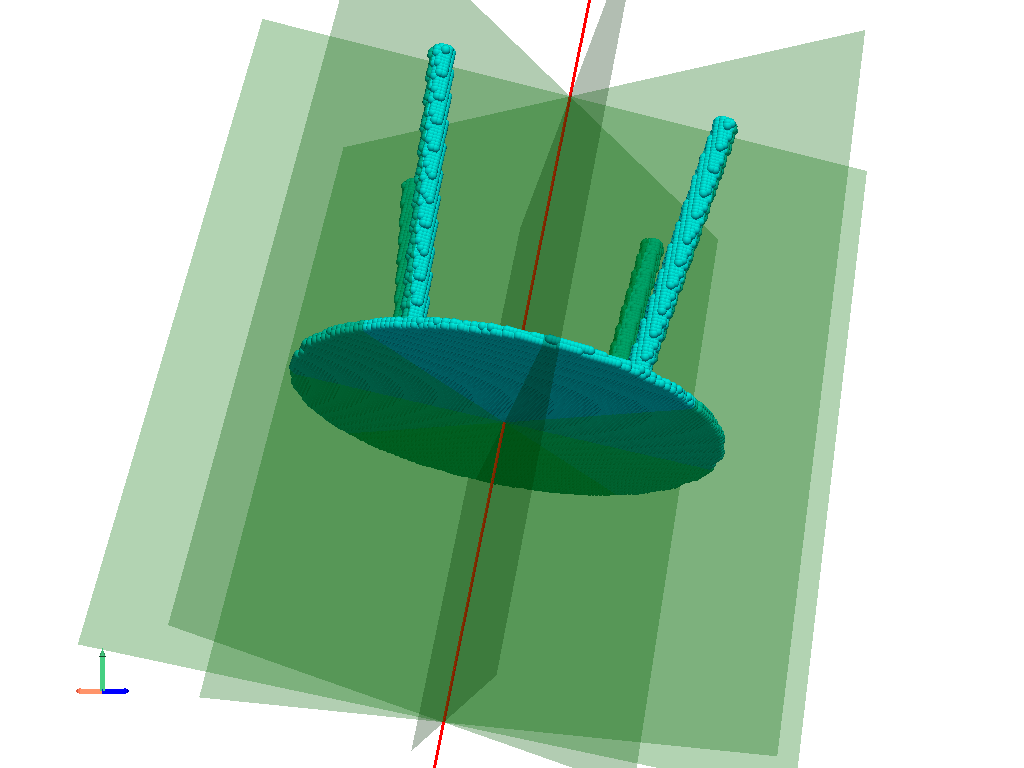}
	}
	\centerline{(a)\hspace{2.5cm}(b)\hspace{2.5cm} (c) }
	\caption{n-plane symmetry case: (a) table model, (b) the rotated table with added noise, (c) the $n$-tuples computed from moments }
	\label{fig:table_multiplane}
\end{figure}


This section addresses objects exhibiting symmetry concerning multiple planes and their rotation around a central axis. Specifically, these objects possess $b$ symmetry planes. An illustrative example is provided by the model of a four-legged table depicted in Fig.~\ref{fig:table_multiplane}. This table exhibits four distinct symmetry planes: two located between the legs and two intersecting them.

This issue can be addressed through iterative approximation methods, as demonstrated in the works of \cite{Funk2017,ecins2017detecting,Nagar2020,Cicconet2017}. However, the accuracy and ability of iterative solutions to detect all planes depend on their initialization. Therefore, by initializing the problem close to the optimal solution, we stabilize the convergence and enhance performance.



To illustrate how the $n$-tuples description can be integrated with iterative methods, we employed the method proposed in \cite{ecins2017detecting} to solve the problem of $b$ symmetrical planes, augmented by our proposed method presented in sections 3 and 4. In this experiment, the estimated axial symmetry from moment $n$-tuples serves as a suitable candidate for initializing the search for $b$-planes of symmetry.

The residual function, as proposed in \cite{ecins2017detecting}, and its gradient are defined using a single random variable $\theta$, which rotates the symmetry plane $\vec{S}$ as follows,

\vspace*{-3mm}
\[
	e= \left(\mathbf{R}^{\left[\theta\vec{\bf \upsilon} \cdot \vec{S}\right]} {\bf F}\left(\mathbf{R}^{\left[\theta\vec{\bf \upsilon} \cdot \vec{S}\right]}\right)^{\textstyle{-1}} \mathbf{x}_u  - \mathbf{x}_l\right)\vec{{\bf n}}_l \quad,
\]
\begin{figure}[t]
    \vspace{-0.75cm}
	\centerline{
		\includegraphics[width=.4\columnwidth]{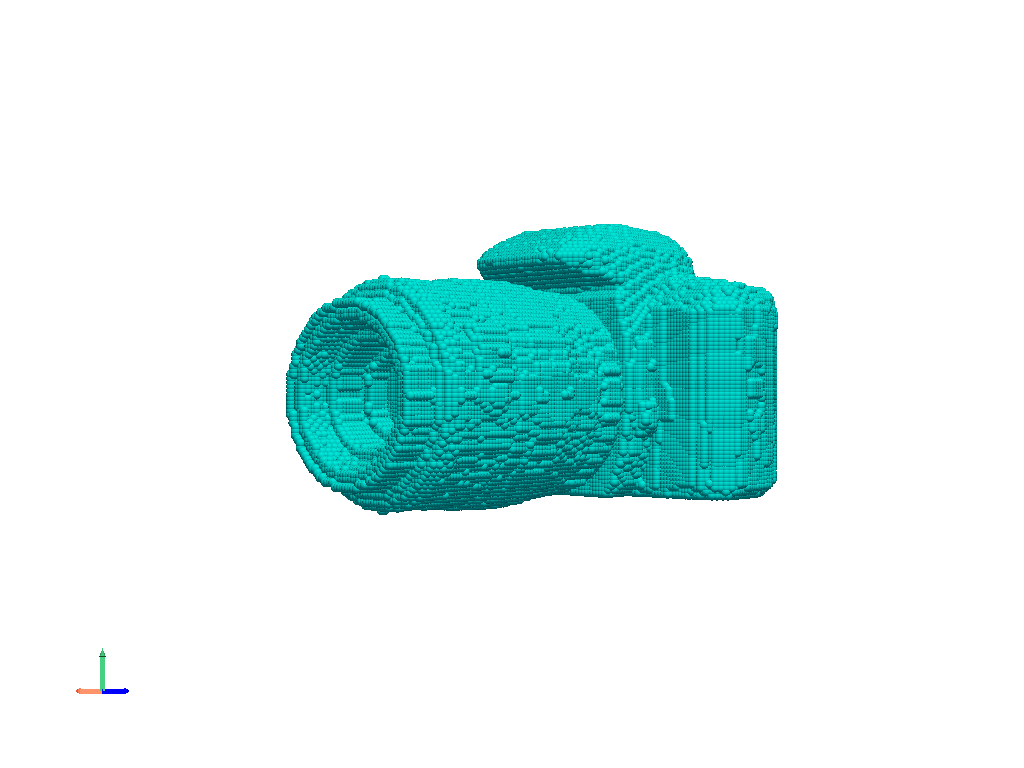}
		\includegraphics[width=.4\columnwidth]{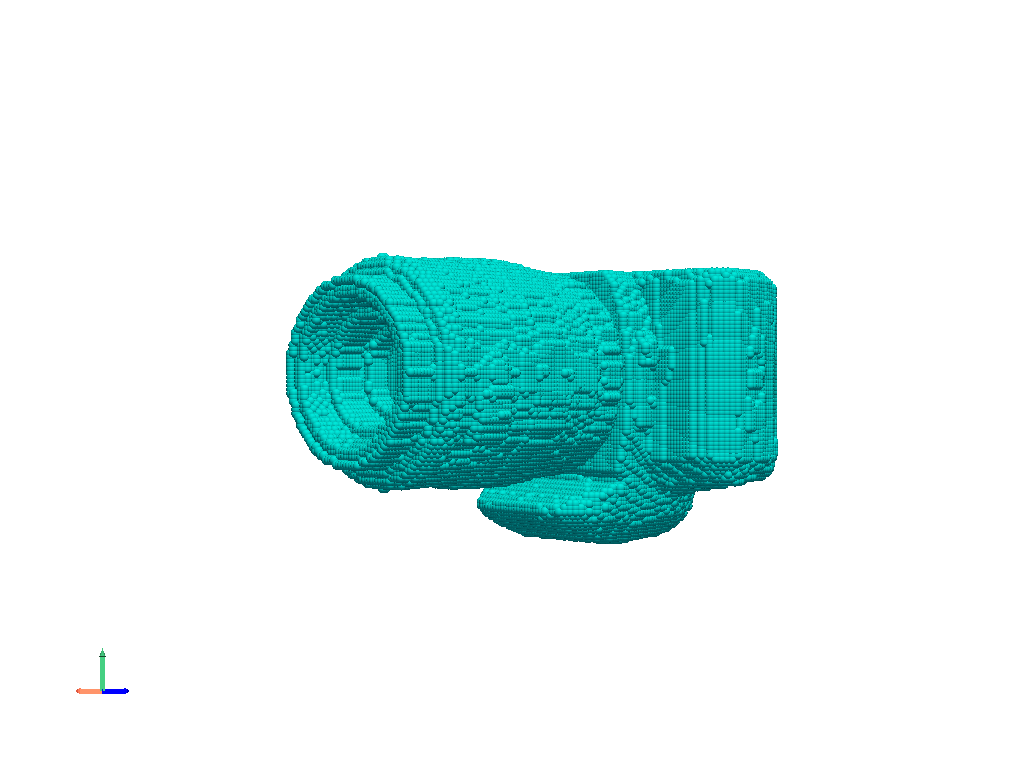}
  }
    \vspace{-0.5cm}
\centerline{(a)\hspace{3.2cm}(b) }
  \centerline{
		\includegraphics[width=.4\columnwidth]{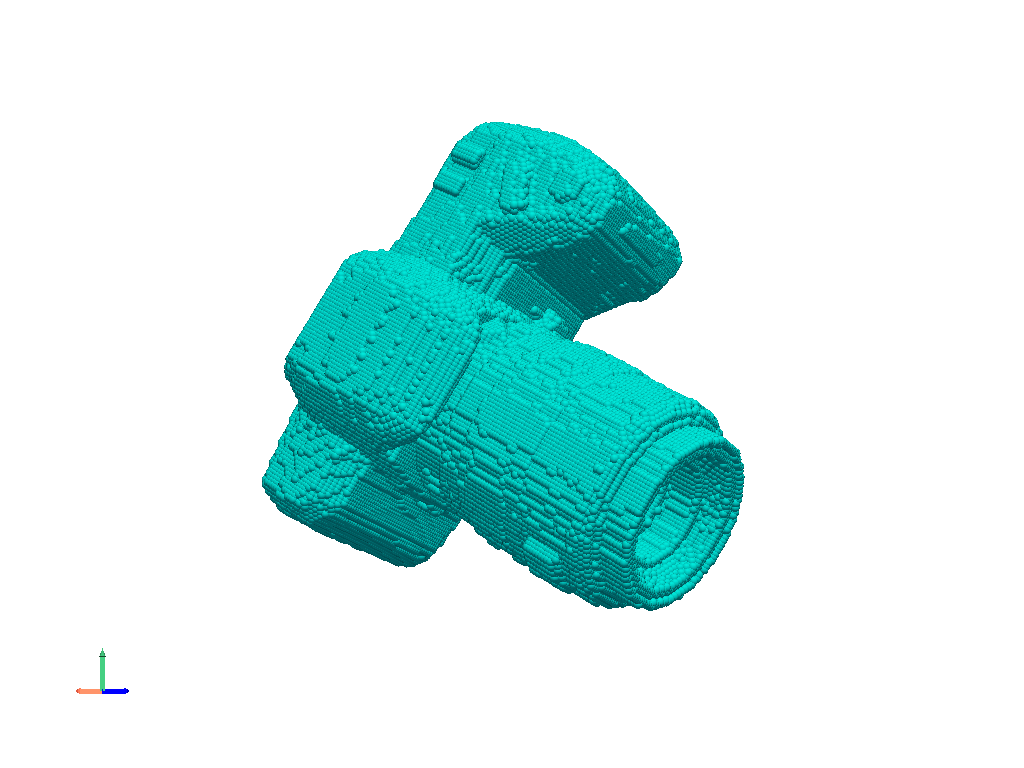} 			
		\includegraphics[width=.4\columnwidth]{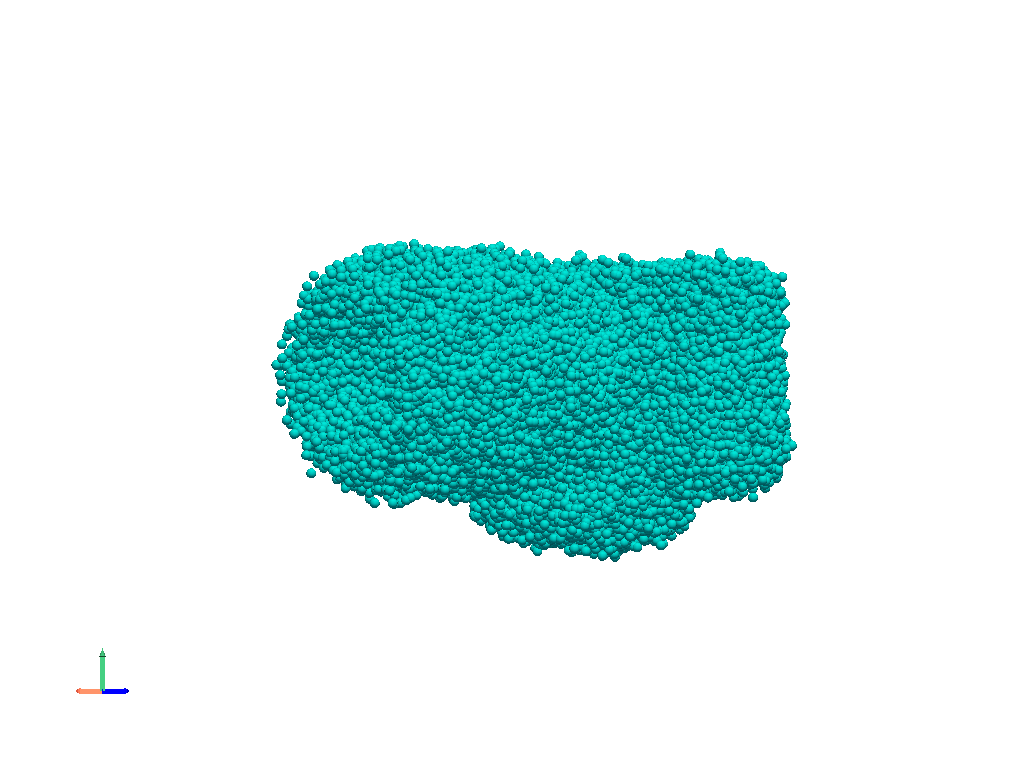}
  }
    \vspace{-0.5cm}
	\centerline{(c) \hspace{3.2cm}  (d) }
	\caption{Reflection estimation using camera model: (a)  model, (b) after reflection, (c) after reflection then rotation, (d) ,(e) noisy model}
	\label{fig:reflectionhand}
\end{figure}

\noindent where $\mathbf{x}_u$ and $\mathbf{x}_l$ are respectively upper and lower 3D points with respect to the symmetry plane defined by $\vec{S}$. And
$\vec{{\bf n}}_l$ is the orientation of the lower point $\mathbf{x}_l$. Remark, that $\mathbf{R}^{\left[\cdot\right]}$ function represent the argument of an orthogonal transformation SO(3) and ${\bf F}$ is a diagonal matrix defined by the vector $\left[\quad 1, \quad 1, \quad -1\right]$. This formulation decrements the number of iteration to converge to a correct solution, beside with the improvement of computational cost per iteration.
In the case of Fig.~\ref{fig:table_multiplane}, the $n$-tuples iterative method recover all 4 planes of symmetry after 3 iteration.
While the baseline iterative method succeeded to get 3 planes of symmetry after 10 iterations (Fig.~\ref{fig:table_multiplane} (b)).

\subsection{Reflection estimation}
\begin{figure*}[t]
	\centerline{
		\includegraphics[width=.12\textwidth]{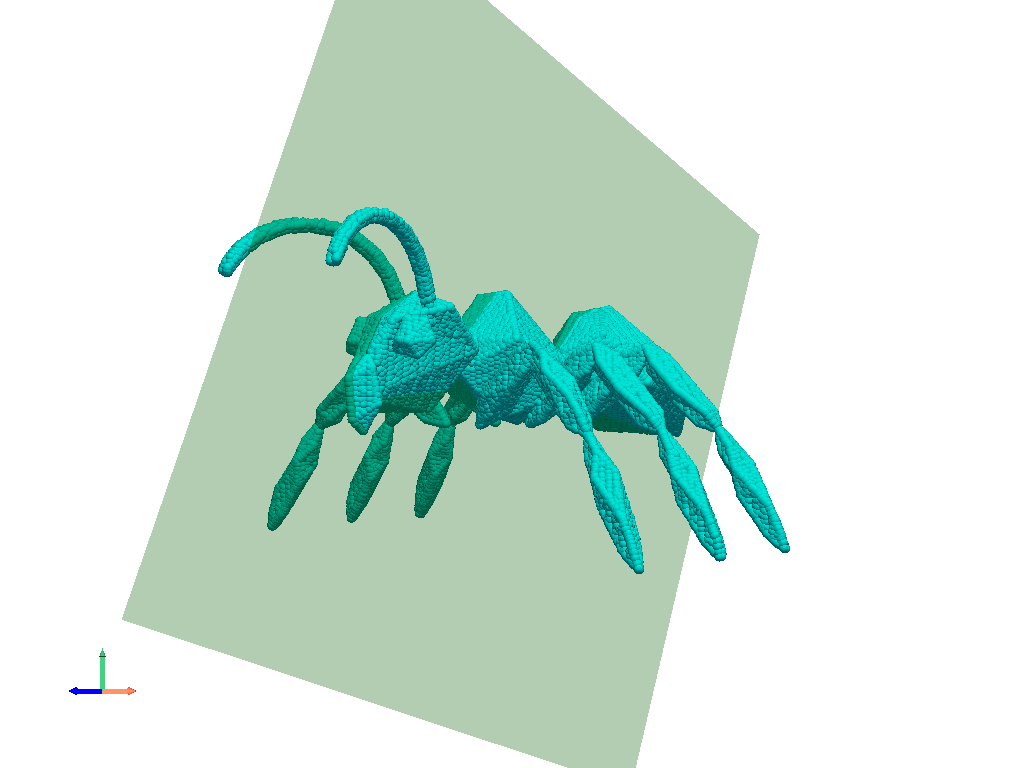}
		\includegraphics[width=0.12\textwidth]{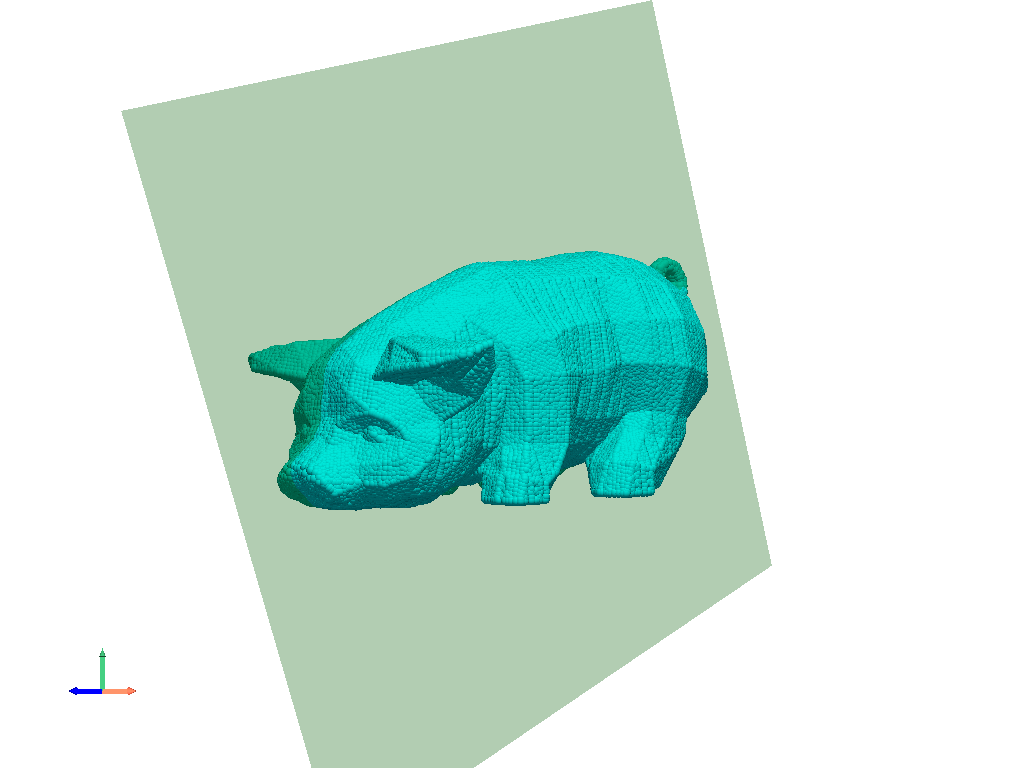}
		\includegraphics[width=0.12\textwidth]{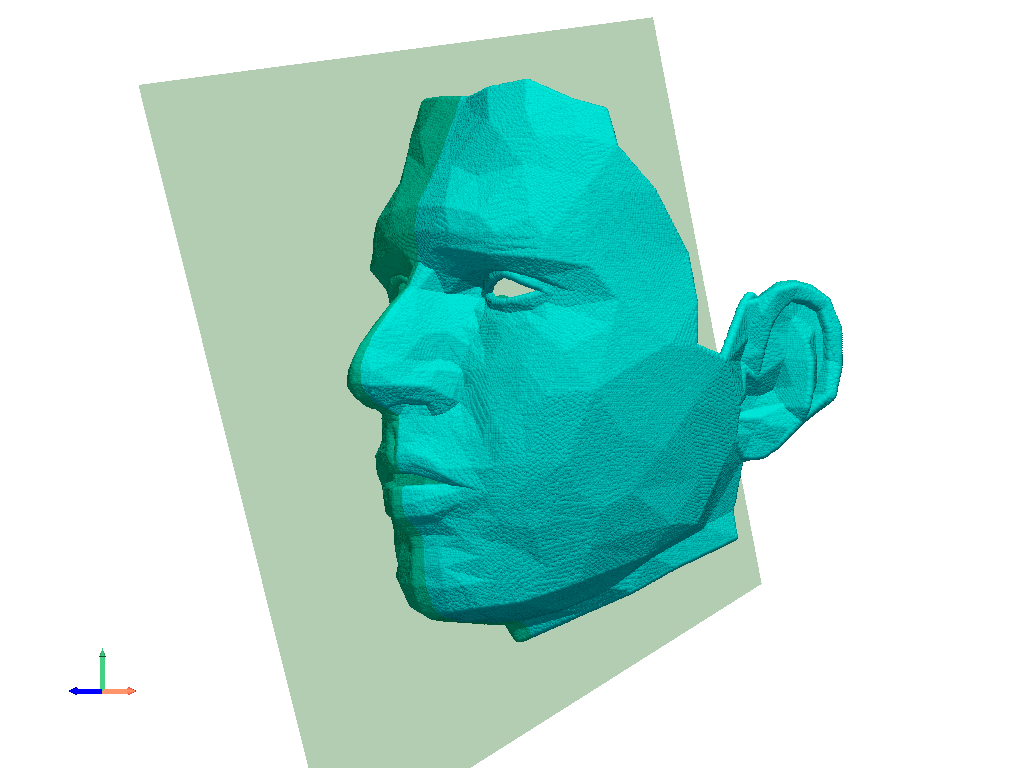}
		\includegraphics[width=0.12\textwidth]{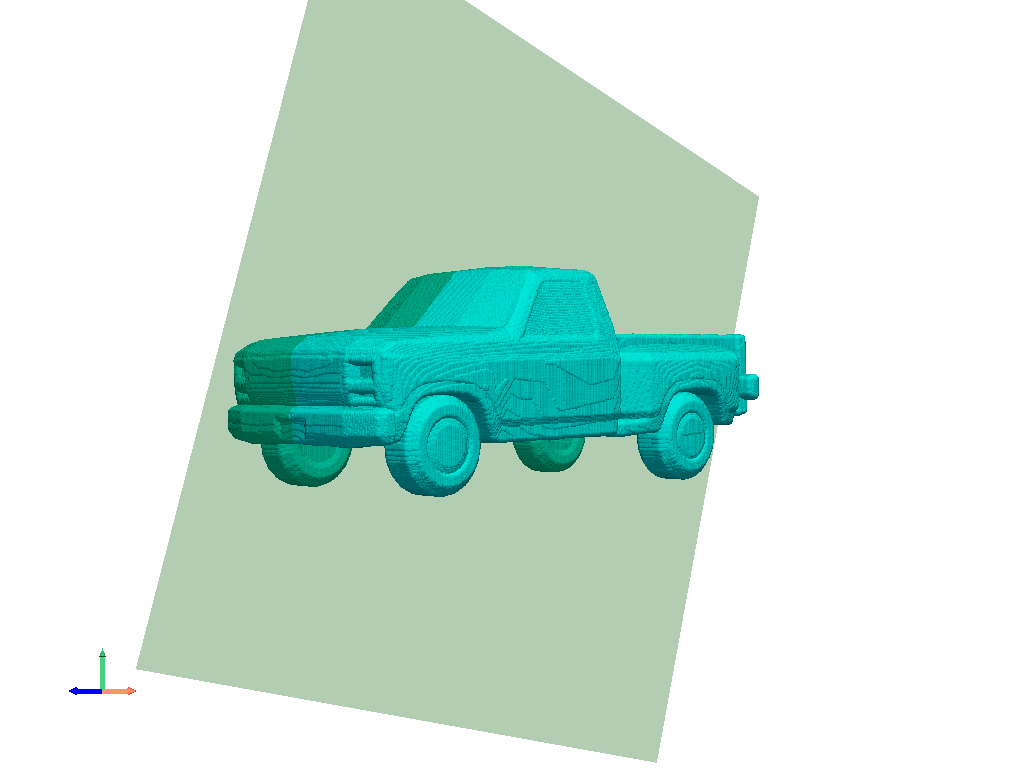}
	   \includegraphics[width=0.12\textwidth]{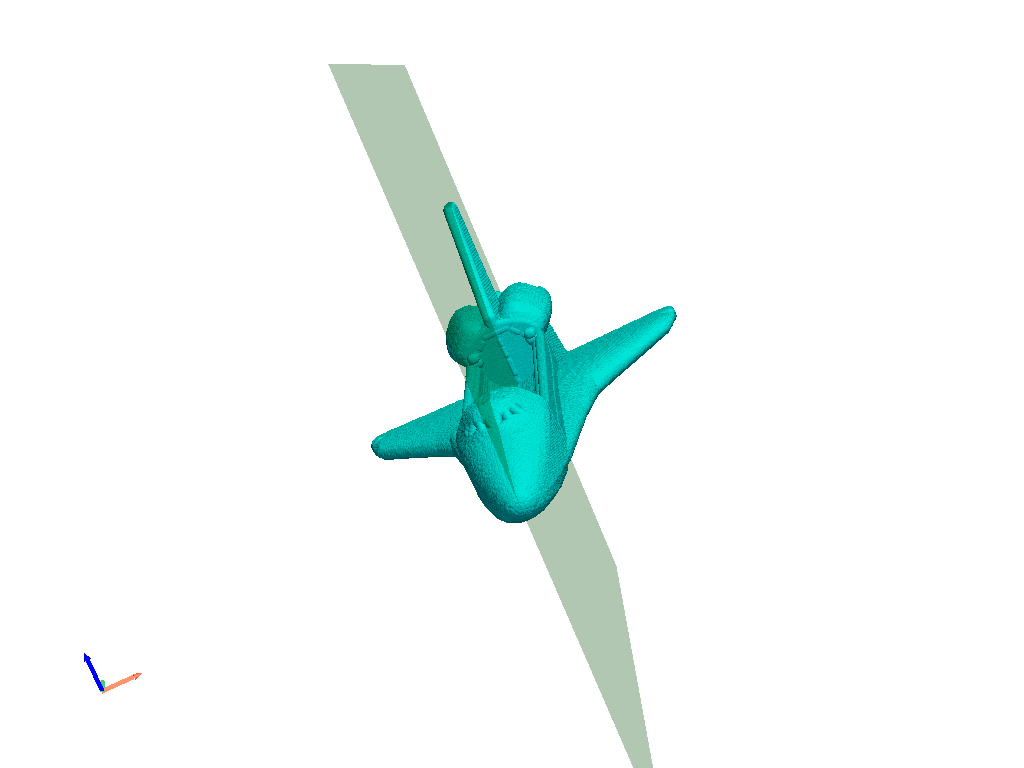}
		\includegraphics[width=0.12\textwidth]{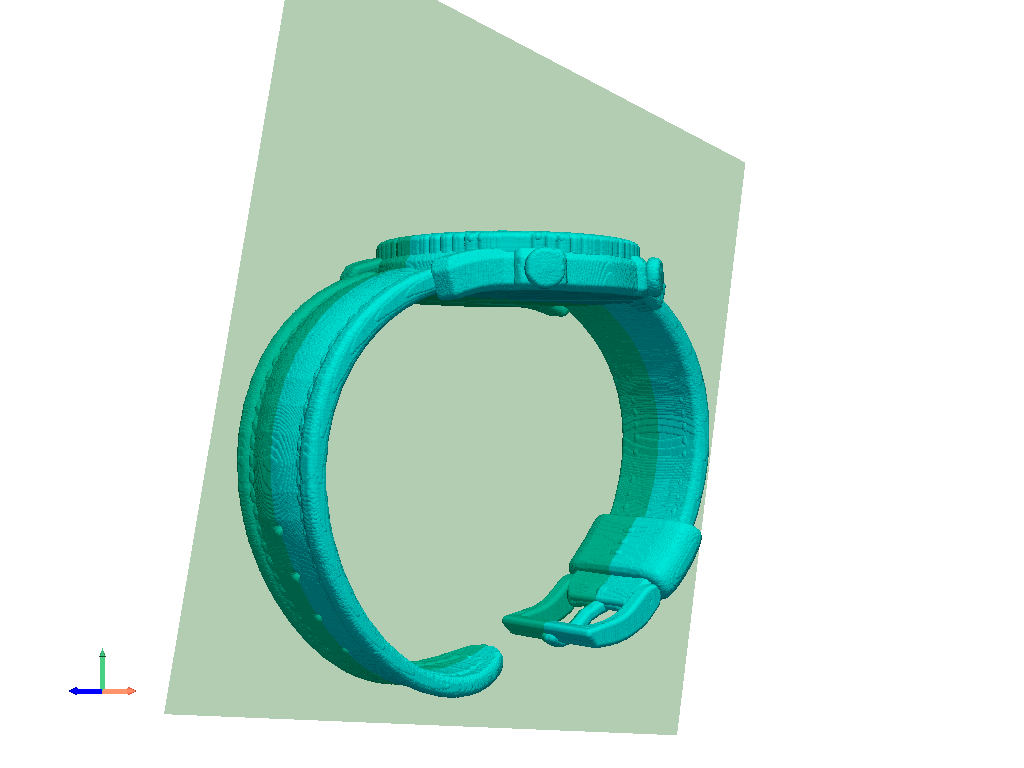}
		\includegraphics[width=0.12\textwidth]{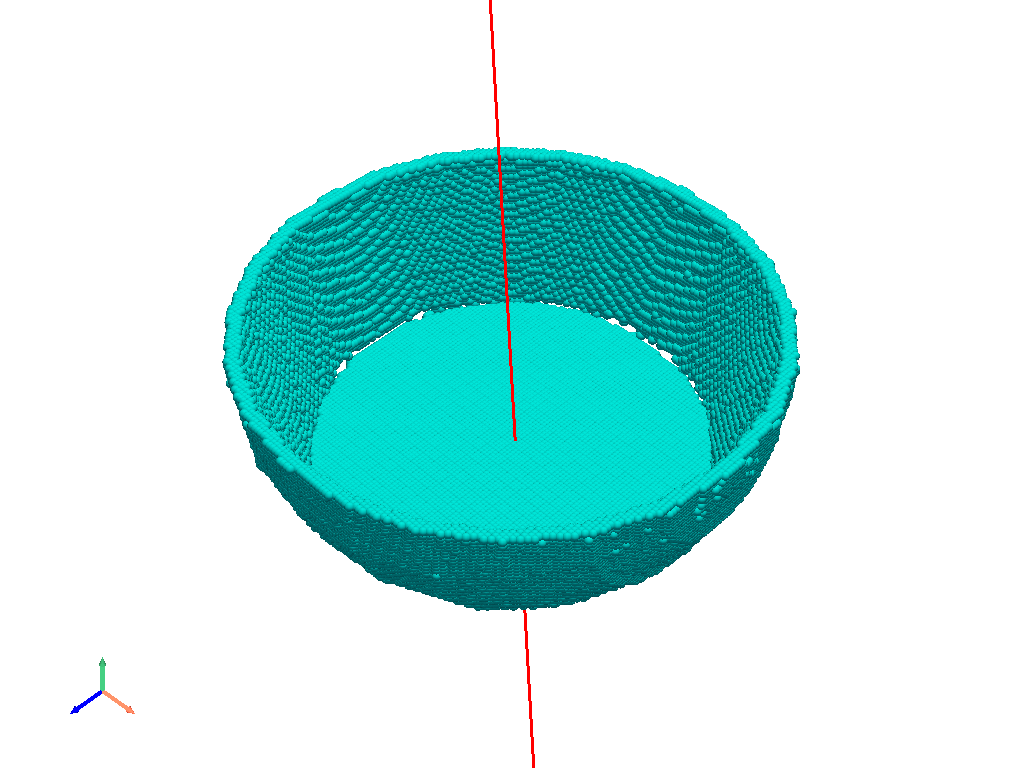}   }
	\centerline{
  		\includegraphics[width=0.12\textwidth]{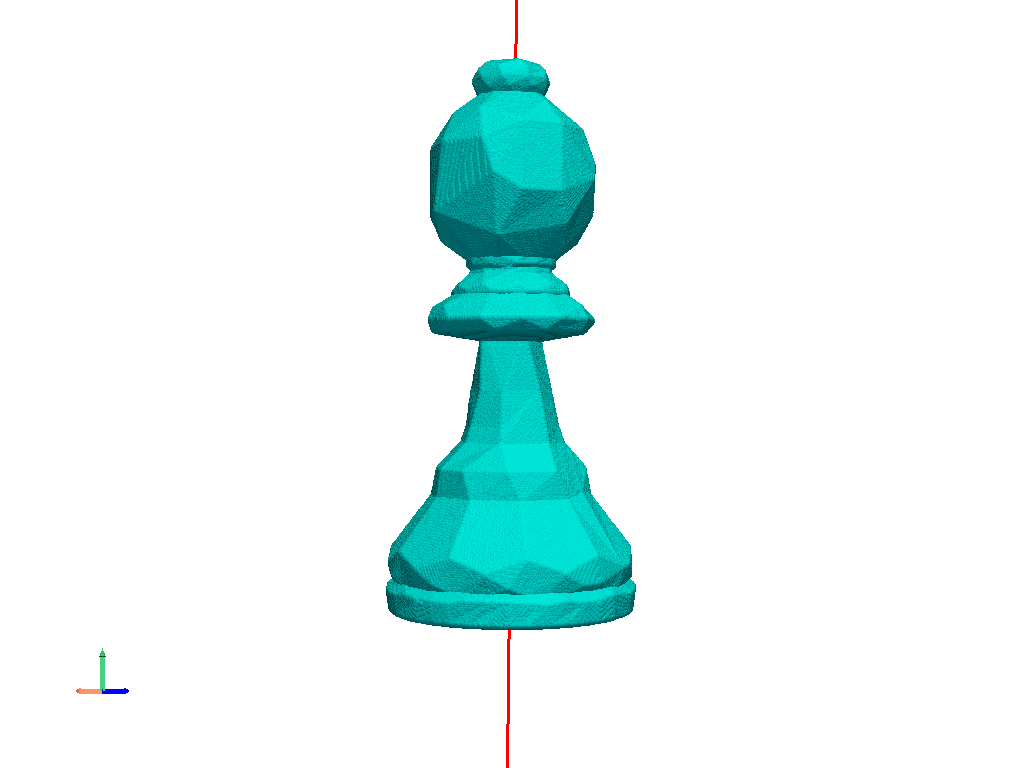}
		\includegraphics[width=0.12\textwidth]{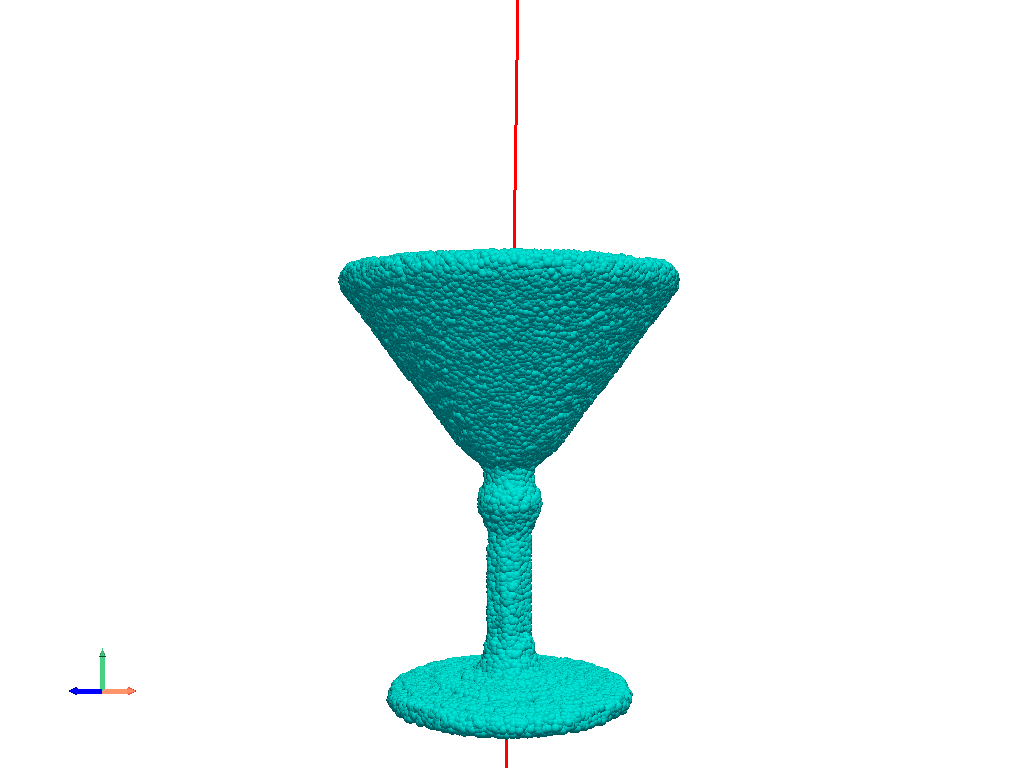}
     \includegraphics[width=0.12\textwidth]{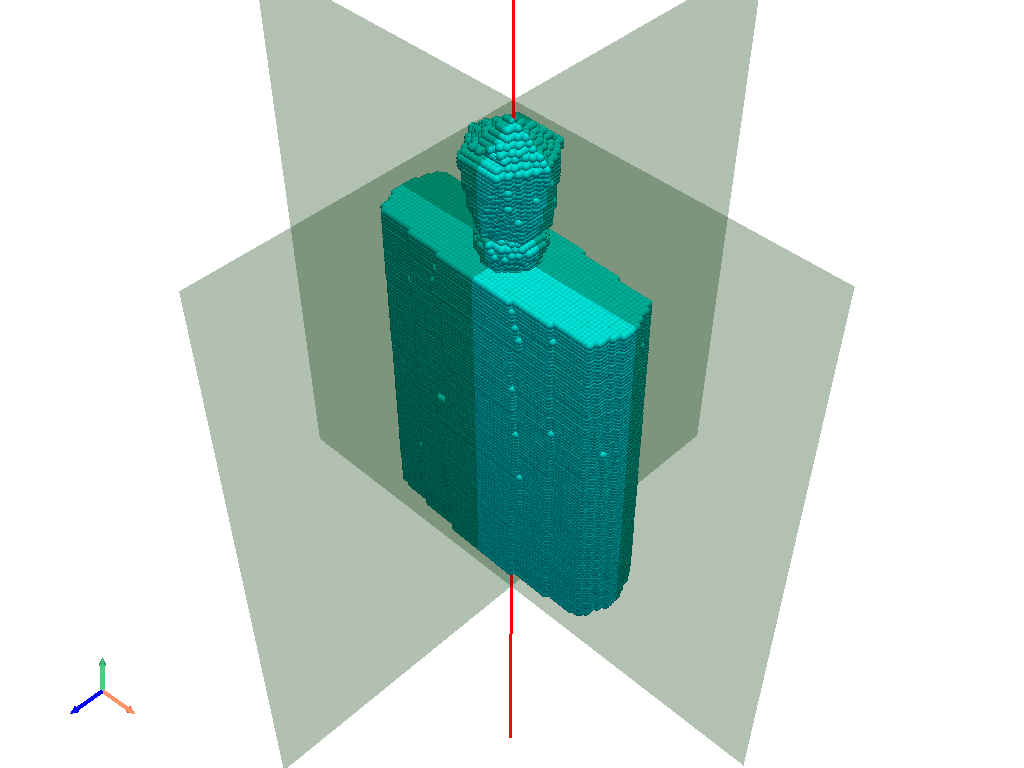}
		\includegraphics[width=0.12\textwidth]{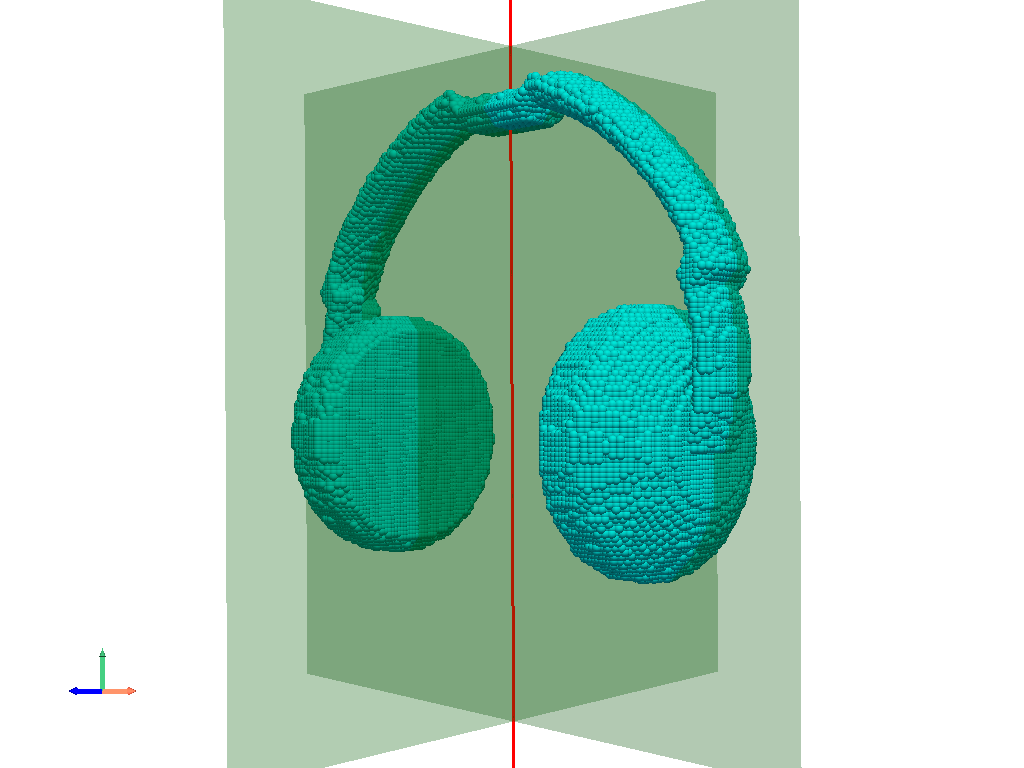}
		\includegraphics[width=0.12\textwidth]{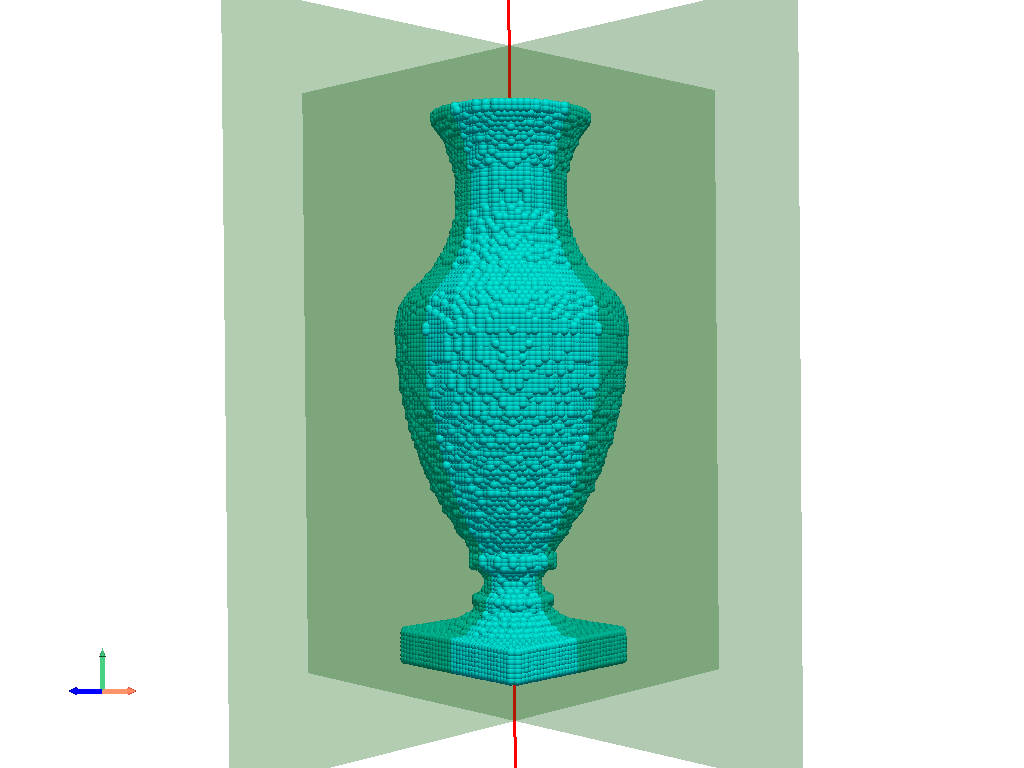}
		\includegraphics[width=0.12\textwidth]{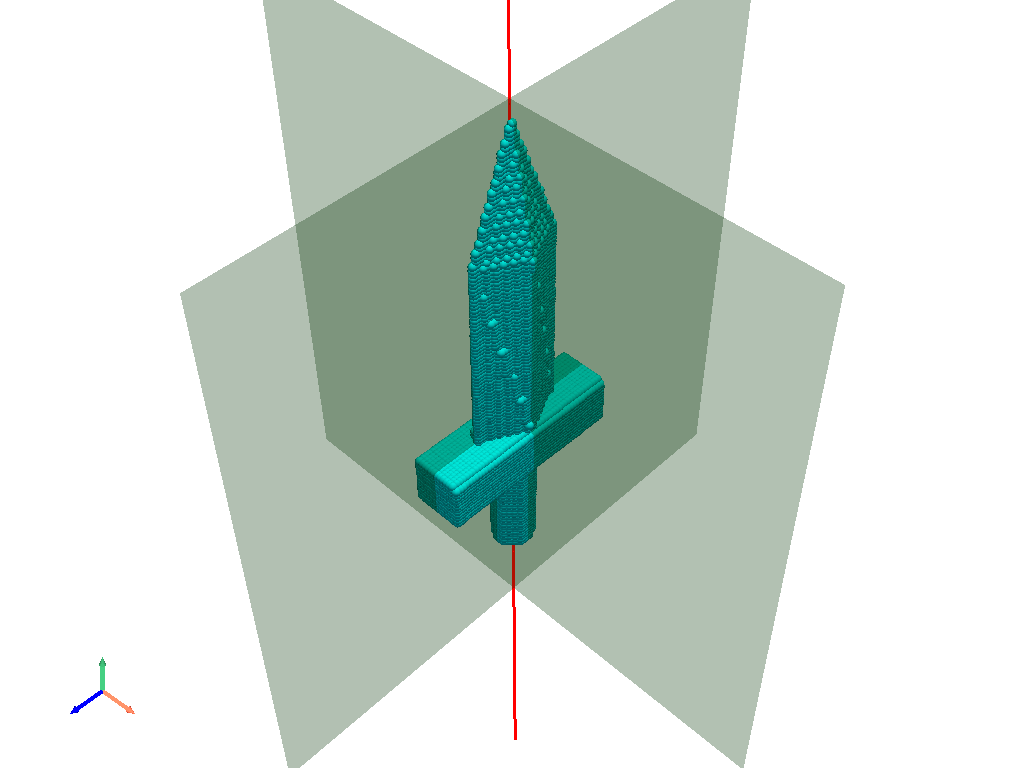}
		\includegraphics[width=0.12\textwidth]{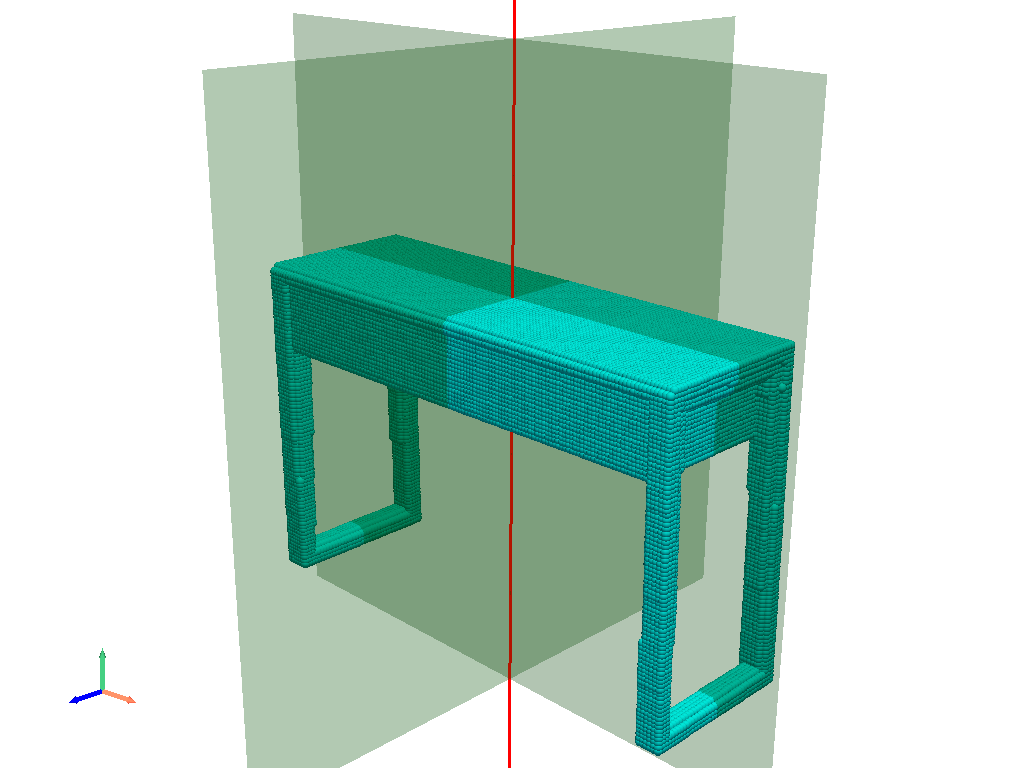}
   }
	\caption{Examples of symmetry with respect to a plane and to an axis\label{otherexamples}}
\end{figure*}

In this last experiment, we validate orthogonal transformation estimation using a non-symmetrical object, particularly the used model is a reflex camera (Fig.~\ref{fig:reflectionhand}).
We consider three different poses in order to conduct the following experiment.
The reference one, is aligned to the world frame, the second one is generated by applying a reflection transformation $\mathbf{F}$ defined in eq.~({\ref{FHANDS}}), and the third one is obtained by applying a random rotation transformation $\mathbf{R}$ after applying $\mathbf{F}$ to original pose eq.~({\ref{FRHANDS}}) (Fig.~\ref{fig:reflectionhand} (c)).
As it can be observed, moments $n$-tuples recover the SO(3) and reflection transformations in absence of noise.
These properties it is conserved no matters if the object is symmetrical or not.
\begin{equation}
	\label{FHANDS}
	{\bf F}=\left[
		\begin{array}{ccc}
			1 & 0  & 0 \\
			0 & -1 & 0 \\
			0 & 0  & 1
		\end{array}\right]=\widehat{{\bf F}}
\end{equation}
\begin{equation}
	\label{FRHANDS}
	{\bf R}_{\bf F}=\left[
		\begin{array}{ccc}
			-0.3085 & 0.2118 & 0.9273  \\
			-0.8599 & 0.3546 & -0.3671 \\
			0.4066  & 0.9106 & -0.0727
		\end{array}\right]=\widehat{{\bf R}_{\bf F}}
\end{equation}
Finally, for the same experiments, we add Gaussian noise $\mathcal{N}\left(0~|~5\cdot 10^3\right)$ (Fig.~\ref{fig:reflectionhand} (d)) in order to demonstrate the performance of our method under noise conditions.
As it is shown in eq.~(\ref{FHANDS_noise}) and eq.~(\ref{FRHANDS_noise}) in both cases recovered transformations are close to the optimal solution.
Concretely, both have a similar deviation of $1 \cdot 10^{-2}$ using mean squared error of the differences.
\begin{equation}
	\label{FHANDS_noise}
	{\bf F}=\left[
		\begin{array}{ccc}
			1 & 0  & 0 \\
			0 & -1 & 0 \\
			0 & 0  & 1
		\end{array}\right] \approx \widehat{{\bf F}}
\end{equation}
\begin{equation}
	\label{FRHANDS_noise}
 \hspace{-0.8cm}
	{\bf R}_{\bf F}=\left[
		\begin{array}{ccc}
			0.466 & 0.546  & 0.695  \\
			0.846 & -0.048 & -0.530 \\
			0.256 & -0.835 & 0.485
		\end{array}\right]\approx \widehat{{\bf R}}_{\bf F}
\end{equation}
As final validation results,  Fig.~\ref{otherexamples} shows  the application of the method using moments $n$-tuples under different conditions: reflection; axial; and $b$-plane symmetrical.
It can be seen that the obtained results are consistent with the discussed above.
Hence, the effectiveness of the proposed scheme. The source code and $28$ object example can be downloaded from the following URL\footnote{ \url{https://drive.google.com/file/d/1ija_bMnA3I0SXmdGsRqzAl_EbZFc5tcw/view?usp=sharing}}.

\vspace*{-4mm}
\section{Conclusion and future works}
\vspace*{-2mm}

This paper has proposed unique measures called moments $n$-tuples to detect symmetries and to estimate orthogonal transformations, including reflections in $n$-dimensional space.
Here, we focus the analysis concretely in 2 and 3-dimensional cases, but it is extensible to $n$ dimensional problems.
We show that those two properties, symmetry and orthogonal transformation, are linked in our proposed moments $n$-tuples space.
Besides, note that symmetry affects rotation observability since it causes ambiguity on its estimation.
This can be seen through the distribution of the $n$-tuples in the space.
The validation results show that the proposed method is able to detect symmetry even for object that are not completely symmetrical. 
They also show a low computation cost. 
Future works will be devoted to extend the obtained results to affine transformation estimation instead of only orthogonal ones.
In fact, if a continuous distribution in $\mathbb{R}^2$ undergoes an affine transformation, we can show for instance that the doublet, defined as follows, \[
	{\bf x_{a}}=\frac{1}{m_{00}^4}\left[\begin{array}{c}
			m_{02}m_{30 }- 2m_{11}m_{21 }+ m_{12}m_{20} \\
			m_{02}m_{21 }+ m_{03}m_{20 }- 2m_{11}m_{12 }
		\end{array}
		\right]
\] will undergo the same transformation as the source distribution. 
We will be concerned with a general scheme for determining affine $n$-tuples in $n$-dimensional space.



\bibliographystyle{elsarticle-harv} 
\bibliography{bibfile}





\end{document}